\documentclass[conference]{IEEEtran}
\IEEEoverridecommandlockouts
\usepackage{comment}
\usepackage{amsmath,amsfonts}
\usepackage{array}
\usepackage[caption=false,font=normalsize,labelfont=sf,textfont=sf]{subfig}
\usepackage{textcomp}
\usepackage{stfloats}
\usepackage{url}
\usepackage{verbatim}
\usepackage{graphicx}
\usepackage{array}
\usepackage{caption} 
\usepackage{xcolor}
\usepackage{multirow}
\usepackage[ruled, lined, linesnumbered, longend]{algorithm2e}
\usepackage{soul} 

\def\BibTeX{{\rm B\kern-.05em{\sc i\kern-.025em b}\kern-.08em
    T\kern-.1667em\lower.7ex\hbox{E}\kern-.125emX}}
    
\SetKwInput{KwOutput}{Output}   
\SetKwInput{KwInput}{Input}

\newcommand{\rev}[1]{{\color{black}{}\color{black}{#1}}{\color{black}}}

\newcommand{\AD}[1]{\textcolor{blue}{#1}}

\begin{document}

\title{(Mis)leading the COVID-19 vaccination discourse on Twitter: An exploratory study of infodemic around the pandemic}

\author{
\IEEEauthorblockN{Shakshi Sharma}
\IEEEauthorblockA{\textit{Institute of Computer Science} \\
University of Tartu, Estonia\\
shakshi.sharma@ut.ee}
\and
\IEEEauthorblockN{Rajesh Sharma}
\IEEEauthorblockA{\textit{Institute of Computer Science} \\
University of Tartu, Estonia\\
rajesh.sharma@ut.ee}
\and
\IEEEauthorblockN{Anwitaman Datta}
\IEEEauthorblockA{\textit{School of Computer Science and Engineering} \\
Nanyang Technological University, Singapore\\
anwitaman@ntu.edu.sg}
}

\maketitle

\begin{abstract}
    In this work, we collect a moderate-sized representative corpus of tweets (over 200,000) pertaining to COVID-19 vaccination spanning for a period of seven months (September 2020 - March 2021). Following a Transfer Learning approach, we utilize a pre-trained Transformer-based XLNet model to classify tweets as Misleading or Non-Misleading and manually validate the results with random subsets of samples. We leverage this to study and contrast the characteristics of tweets in the corpus that are misleading in nature against non-misleading ones. This exploratory analysis enables us to design features such as sentiments, hashtags, nouns, pronouns which can, in turn, be exploited for classifying tweets as (Non-)Misleading using various Machine Learning (ML) models in an explainable manner. Specifically, several ML models are employed for prediction, with up to 90\% accuracy, with the importance of each feature is explained using SHAP Explainable AI (XAI) tool. While the thrust of this work is principally exploratory in nature in order to obtain insights on the online discourse on COVID-19 vaccination, we conclude the paper by outlining how these insights provide the foundations for a more actionable approach to mitigate misinformation. We have made the curated data as well as the accompanying code available so that the research community at large can reproduce, compare against or build upon this work.
\end{abstract}


\footnote{This is a preprint version of the accepted paper in IEEE Transactions on Computational Social Systems 2022.}

\section{Introduction}\label{sec:Intro1}


\textbf{\textit{“We live in an era of unprecedented scientific breakthroughs and expertise. But we’re also stymied by the forces of misinformation that undermine the true knowledge that is out there.”} -- Dr. Laolu Fayanju} \cite{bosman2021unvaccinated}

The fears, uncertainties, and doubts (FUD) around COVID-19 and the available vaccines have been amplified by the ongoing discourse on social media - leading to an infodemic (a portmanteau of information and epidemic, referring to the spread of possibly accurate and inaccurate information about a disease, itself spreading like an epidemic). \rev{The infodemic has been fuelled by prolific misinformation spreaders, e.g., the `disinformation dozen' \cite{disinfo2021dozen}, who may have special interests and agendas in doing so. \cite{Machingaidze2017hesit} studied vaccine hesitancy in several low and middle-income countries (LMIC) and compared that with the US and Russia, which were some of the countries at the forefront of COVID-19 vaccine research. The average vaccine acceptance rate in LMICs was reported to be 80.3\%, compared to 64.6\% in the United States and 30.4\% in Russia. Such vaccination hesitancy has made the impact of a harsh pandemic much worse than necessary, e.g., a study estimated a factor of up to 8 more fatalities \cite{report43} attributed to this.}



\rev{Given the pivotal role of social media both as a medium of propagation of misleading information, as well as an easy source to probe the prevailing sentiments at scale, we chose to use openly accessible COVID-19 vaccine-related tweets to drive our research. Unlike traditional surveys, for example, \cite{bosman2021unvaccinated} which are able to collect and analyze detailed demographic breakdowns such as age, race, gender, education, income, geography, etc., our approach is not granular. However, the advantages of sensing information from open social media are its sheer scale and the possibility to carry out continuous and real-time monitoring once the data pipeline is set up\footnote{This work presents only a retrospective analysis.}.}

\rev{Firstly, we wanted to establish an understanding of what are the main topics of discussion and contentions around COVID-19 vaccination in order to identify the drivers of vaccine hesitancy. This exploratory analysis can provide actionable insights to policy-makers on how to approach and address vaccine hesitancy.}

\rev{Second, we wanted to pinpoint misleading tweets promptly and in an automated yet explainable manner. Being able to recognize misleading tweets at scale is useful to identify when and which specific misinformation is gaining traction. This could be useful in many ways. Public health policy-makers can use the information to prioritize their combat against a particular sub-genre of misinformation in a timely fashion. It can aid the governance of the social media platform, e.g., by censoring or tagging misleading tweets - as might be applicable as per terms of usage and regulatory requirements. Outside the scope of this work, it could also help determine suitable counter-points, and timely automated  rebuttals of misleading information (this is a follow-up study we are currently carrying out). The explainability aspect of the classification process proximately helps in validating the quality of the classification. In the longer run (outside the scope of the current paper), it could help decipher temporal shifts in the behavior of misinformation propagators (which could help update and keep the ML models efficacious) and also generalize the approach beyond the current topic.}

Accordingly, the key highlights and contributions of the current work are as follows:

\begin{enumerate}
    \item \textbf{Data}  (following FAIR principle, i.e., findable, accessible, interoperable, and reusable)\textbf{:} From an initial set of over 200,000 tweets which we originally collected, we curated a representative denoised collection of 114,635 tweets related to COVID-19 vaccine, along with two mutually exclusive randomly sampled subsets of 1,500 and 1,000 tweets manually labeled in terms of whether they are \textit{Misleading} or \textit{Non-Misleading}. We have provided the \textit{dataset}\footnote{\url{https://researchdata.ntu.edu.sg/dataset.xhtml?persistentId=doi:10.21979/N9/QMLIJQ}} \rev{comprising separately the manually labeled data used for training and testing, as well as the results of algorithmic prediction over a larger corpus of data, while adhering to Twitter's content redistribution policy}\footnote{\url{https://developer.twitter.com/en/developer-terms/agreement-and-policy}}. The accompanying \textit{source code} has also been open-sourced\footnote{https://github.com/shakshi12/Misleading-Covid-vaccination-Tweets}.
    
\item \textbf{Exploratory analysis:}
   We analyze the dataset across three dimensions: (i) \textit{Language Exploration} utilizing syntactic structure and the principal themes involved across \textit{Misleading} and \textit{Non-Misleading} tweets, (ii) \textit{Opinion Study} leveraging the sentiments and emotions of both types of tweets, (iii) \textit{Effect of Visibility} analyzing meta-data of the tweets. These provide us insights to distinguish \textit{Non-Misleading} from \textit{Misleading} tweets.
      
\item \textbf{Classification \& explainability:} Aforementioned analysis aided the identification of potential features which could be explicitly leveraged to classify tweets to determine whether they are \textit{Misleading} or not. This explicit approach was compared against a black-box model (XLNet). \rev{We also looked into the mutual consistency of these approaches. The explainability aspect} of our approach reinforces the credibility of the classifiers. The efficacy as well as marginal contributions of a subset of features were explored empirically for further validation.

\end{enumerate}

\section{Related Work}\label{sec:background}
In this work, we broadly discuss the literature from two perspectives, namely, online discourse on COVID-19 vaccinations and misinformation.

\noindent{\textbf{Online Discourse on COVID-19 Vaccinations}}
The COVID-19 vaccination debate \cite{wu2021characterizing} has been carried out on social media sites such as Reddit. The vast majority of Reddit thread comments collected are about conspiracy theories related to COVID-19 vaccines. On Twitter, utilizing temporal dimension \cite{malagoli2021look}, 12 million tweets in two months were released. Moreover, primary themes and user-level engagements around the pandemic were discussed. After the initiation of the Italian vaccination campaign, researchers \cite{pierri2021vaccinitaly} have been monitoring the online conversations of Italian Twitter users. They have discovered that there is a larger occurrence of high-credibility information despite the sharing of low-credibility information.
The research \cite{deverna2021covaxxy} included a dashboard with statistics on the COVID-19 vaccination by collecting and analyzing English tweets.
Another research \cite{chen2020tracking} gathered almost 123 million multilingual tweets.
The authors \cite{malagoli2021look} collected the COVID-19 vaccinations' tweets for two months. Analyzed the tweets from two perspectives: user engagement and content properties in a temporal manner. Besides, COVID-19 fact-check dataset has also been collected and analyzed \cite{sharma2022facov}.

\noindent{\textbf{Misinformation}}
Misinformation is an umbrella term used for fake news, rumor, misleading, etc. The research has been performed on rumors \cite{sharma2021identifying, butt2021goes}, fake news \cite{dhawan2022game, mayank2021deap}, even on misinformation \cite{jagtap2021misinformation}. However, much less focus is on the misleading misinformation.
For instance, the authors \cite{ferrara2020covid} focus on the influence of bots in disseminating COVID-19 conspiracy claims. 
The degree of misinformation surrounding the COVID-19 epidemic has also been explored \cite{kouzy2020coronavirus}.
Besides text, the authors \cite{ma2017multimodal} used a multi-modal discourse analysis technique to evaluate textual and visual information inside a public anti-vaccine Facebook page.

The role of Twitter in misinformation during COVID-19 \cite{rosenberg2020twitter} is a concern.
Specifically, \cite{malagoli2021look} is one of the closest to our study, where COVID-19 vaccination tweets are studied. Previous studies lacked an in-depth analysis of the COVID-19 vaccination data from a variety of perspectives.
In the COVID-19 vaccination, for example, there are no significant themes presented in relation to various sentiments and emotions. Moreover, we bring the explainability element to our COVID-19 classification model, which, to our knowledge, has not been investigated on such a dataset. 
To summarize, we employed the transfer learning technique after collecting the tweets, which were then categorized as \textit{Misleading} or \textit{Non-Misleading}. Next, we used NLP-based approaches to analyze, for instance, emotions with respect to \textit{Misleading} and \textit{Non-Misleading} tweets. In addition, we use machine learning techniques to determine if these NLP-based extracted features are adequate to discriminate between \textit{Misleading} and \textit{Non-Misleading} tweets in an explainable manner.

\section{Dataset}\label{sec:dataset}
The data collection and annotation process are covered in this section. The curated dataset (link provided in the Introduction) comprises two separate files, one containing manual labels used for training and testing, and another larger set with algorithmically predicted labels.

\subsection{Data Collection}
The first news of the world's COVID-19 vaccine registration\footnote{https://www.theguardian.com/society/2020/apr/18/coronavirus-vaccine-trials-could-be-completed-by-mid-august} in August 2020, as well as Trump's order to carry out vaccination even before it had been thoroughly tested and approved\footnote{https://www.bbc.com/news/world-us-canada-53899908}, signaled a vaccine rush.
Naturally, this amplified the discussions around Covid-19 vaccines both offline and online. We examine the type of discussions about COVID-19 vaccination which spread over Twitter since Twitter provides an easy to mine source of information representing all the important narratives. We collected tweets related to COVID-19 vaccination from September 2020 to March 2021. 
Most countries had begun vaccine rollouts\footnote{https://www.cnbc.com/2021/03/25/covid-live-updates.html} 
as of March 2021. A significant mass of anti-vaxxers remain, nevertheless, gradually, people are less hesitant to get vaccinated\footnote{https://www.ajmc.com/view/a-timeline-of-covid-19-vaccine-developments-in-2021}, and most of the current narratives and discussion points were already formed and matured in the aforementioned period. As such, we focus on the period leading up to large-scale vaccine roll-out, which is critical for studying and understanding the nature of misinformation and its spread.


Given the Twitter API restrictions, collecting an exhaustive dataset was impractical. As such, we aimed for a representative sample instead. We queried the Twitter Streaming API with a variety of relevant keywords:
\textit{vaccine, covidvaccine, chinesevirus, VaccinesWork, NoMasks, antivaxxer, antivaccine, antivax, COVID19, covax}. 
Since there are no precise keywords for the \textit{Misleading} category, we also considered anti-vaccine tweets to cover a broader range of aspects even though not all \textit{Misleading} tweets are anti-vaccine and vice-versa.
This resulted in over 200,000 tweets. After filtering the tweets\footnote{We removed tweets that aren't relevant to COVID-19 or are not in English.}, 
the final dataset used in this study has 114,635 tweets in English.

\subsection{Data Annotation Process}

In order to perform the analysis, we further process and label the tweets as \textit{Misleading} and \textit{Non-Misleading}. 

\textit{Misleading}, a type of misinformation, is defined as fabricating an issue or an individual that is not true \cite{molina2021fake,zubiaga2018detection}. Specifically, after applying standard NLP techniques to clean the tweets, we designate a tweet to be \textit{Misleading} when the content of a tweet deviates from the evidence shared by news media or reputable sources such as the WHO (World Health Organization), and even if it uses facts in parts, it might add connotations to it that encourages vaccine hesitancy. Otherwise, we consider the tweet as \textit{Non-Misleading}.
For instance, the tweet, 
``\textbf{I know its so bloody weird that Gates funded every single vaccine moderna Pfizer Oxford etc. Something my gut says dont have the vaccine.}''
spins a narrative to increase people's distrust. Thus, in this work we consider it as a \textit{Misleading} tweet. 
The partial truth in the \textit{Misleading} tweets could be the result of the widespread COVID-19 vaccination half-truths and myths\footnote{https://edition.cnn.com/2020/12/18/health/myths-covid-vaccine-debunked/index.html}. 
These narratives are designed to prompt vaccine hesitancy \cite{Mejova2020ICWSM} and eventually become the root cause of further \textit{Misleading} information.

Since data annotation is a costly and labor-intensive method, thus, first manual labeling of a sample subset of tweets is performed by human annotators, which is used for model parameter tuning and validation, when we explored the \textbf{Transfer Learning} \cite{pan2009survey} approach to annotate each tweet in the complete dataset as \textit{Misleading} or \textit{Non-Misleading}. To that end, we first took a random sample of 1,500 tweets from the dataset. This sampling was agnostic of the topics these tweets might be about.
We then manually annotated them as either \textit{Misleading} or \textit{Non-Misleading} tweets.
The sample tweets were labeled by three annotators. 
Then, using Fleiss' kappa score \cite{mchugh2012interrater} \textit{(k)} to test the quality of the dataset annotation, an \textbf{inter-annotator} agreement amongst three annotators was performed, producing \textit{k = 0.84}, validating a high quality of agreement.
The resulting labels had the ratio 47.7:52.3 among \textit{Misleading} and \textit{Non-Misleading} tweets.

\rev{
We experimented with the sampling of tweets by selecting various sample sizes and utilizing five-fold cross-validation. Initially, we fine-tuned the pre-trained XLNet transformer model\footnote{https://huggingface.co/transformers/model\_doc/xlnet.html} with a sample size of 100. The results, as can be seen in Table \ref{tbl:sample_size}, are less promising and show the model’s underfitting. Next, we increased the sample size to 500 and observed an improvement over the previous performance. We continued with increasing the sample size. We observe that the performance dropped when the sample size was 2,000. This is because while the training set's accuracy, precision, recall, and other metrics all reached 0.92, the test set's results were noticeably poorer. Consequently, we used 1,500 samples to fine-tune the pre-trained XLNet model. 
In order to demonstrate the stability of the results, we repeated these experiments five times but with different random 
seeds (1, 124, 2012, 2022, 1000). 
}

\begin{table*}[!htbp]
\centering
\caption{Selection of various sample sizes to fine-tune the pre-trained XLNet model. The values in the cells denote the average performance of the five-fold cross-validation along with the standard deviation. ACC, PR, RC, F1, and AUC represent Accuracy, Precision, Recall, F1-Score, and AUC ROC Score, respectively.}
\label{tbl:sample_size}
\begin{tabular}{|c|c|c|c|c|c|}
\hline
\textbf{Sample size} & \textbf{ACC}  & \textbf{PR}   & \textbf{RC}   & \textbf{F1}   & \textbf{AUC} \\ \hline
100                  & 0.25  ± 0.012 & 0.24  ± 0.005 & 0.24  ± 0.022 & 0.24  ± 0.010 & 0.26  ± 0.100    \\ \hline
500                  & 0.42  ± 0.23  & 0.45  ± 0.023 & 0.44  ± 0.012 & 0.45  ± 0.003 & 0.44  ± 0.302    \\ \hline
1000                 & 0.72  ± 0.009 & 0.77  ± 0.120 & 0.77  ± 0.231 & 0.77  ± 0.011 & 0.73  ± 0.003    \\ \hline
1,500                & 0.86  ± 0.001 & 0.86  ± 0.002 & 0.88  ± 0.001 & 0.86  ± 0.111 & 0.86  ± 0.012    \\ \hline
2,000                & 0.68  ± 0.113 & 0.73  ± 0.023 & 0.72  ± 0.013 & 0.72  ± 0.112 & 0.72  ± 0.302    \\ \hline
\end{tabular}%
\end{table*}

After choosing 1,500 as the sample size for the tweets, we then finally fine-tune the XLNet language model using the manually labeled tweets.
XLNet is an extension of the Transformer-XL model and learns the bidirectional contexts. BERT and RoBERTa were also tested. XLNet outperformed all other models in our case. The 1,500 manually annotated sampled tweets were split into an 80:20 train and test set utilizing a five-fold cross validation technique. Specifically, 300 out of 1,500 tweets were used as a test set, and the rest were used as a training set. Table \ref{tbl:validate}, row 1 shows the evaluation metrics calculated on the test set. The results shown are the average value on the five test sets obtained using the five-fold cross validation technique. Subsequently, we used the fine-tuned XLNet model to annotate the entire dataset.
To validate the efficiency of the annotated tweets, we manually verified 1,000 random tweets. This sample was also balanced in terms of the predicted labels. Table \ref{tbl:validate}, row 2 shows the evaluation metrics of this validation. The accuracy for the sample was 0.86, indicating that our model had been well-trained, and we can rely on these labels to get further insights.
\begin{table}[!htbp]
\centering
\caption{Evaluation metrics of the fine-tuned XLNet model. \rev{ACC, PR, RC, F1, and AUC represent Accuracy, Precision, Recall, F1-Score, and AUC ROC Score, respectively.}}
\label{tbl:validate}
\begin{tabular}{|c|c|c|c|c|c|c|}
\hline
\textbf{Experiment} & \textbf{\# of samples} & \textbf{PR} & \textbf{RC} & \textbf{F1} & \textbf{ACC} & \textbf{AUC} \\ \hline
Test set & 300                  & 0.86         & 0.88          & 0.86         & 0.86  & 0.86        \\ \hline
Validate & 1,000                   & 0.88         & 0.88          & 0.89           & 0.88 & 0.88        \\ \hline

\end{tabular}%
\end{table}

In the following sections, we will provide the analysis of the whole dataset, which is divided into a ratio of 70:30 \textit{non-misleading} and \textit{misleading} tweets, respectively. 



\section{Language Exploration}\label{sec:lang}

In this section, we uncover ten distinct Syntactic styles of \textit{Misleading} and \textit{Non-Misleading} tweets. 

\subsection{Uncover Syntax}
After assigning the labels for each tweet (as discussed in Section \textit{Dataset}), we study ten Syntactic aspects (attributes) which were found to be more significant than others in our dataset to distinguish the structural patterns of both types of tweets. 
We used the NLTK library\footnote{https://www.nltk.org/book/ch05.html} to extract the part-of-speech tags. 
First, we visualize the Syntactic distributions of both \textit{Misleading} and \textit{Non-Misleading} tweets. Next, to validate that the difference in both the distributions is indeed significant, we use Kolmogorov Smirnov Test\footnote{https://www.itl.nist.gov/div898/handbook/eda/section3/eda35g.htm}.

\begin{figure*}[ht!]
\begin{tabular}{|c|c|c|c|c|}
\hline
\centering
(a) Nouns & (b) Pronouns & (c) TTR & (d) Stop words & (e) Verbs \\\hline

\subfloat{\includegraphics[width=0.36\columnwidth, height = 3cm, trim={0cm 0cm 0cm 0cm},clip]{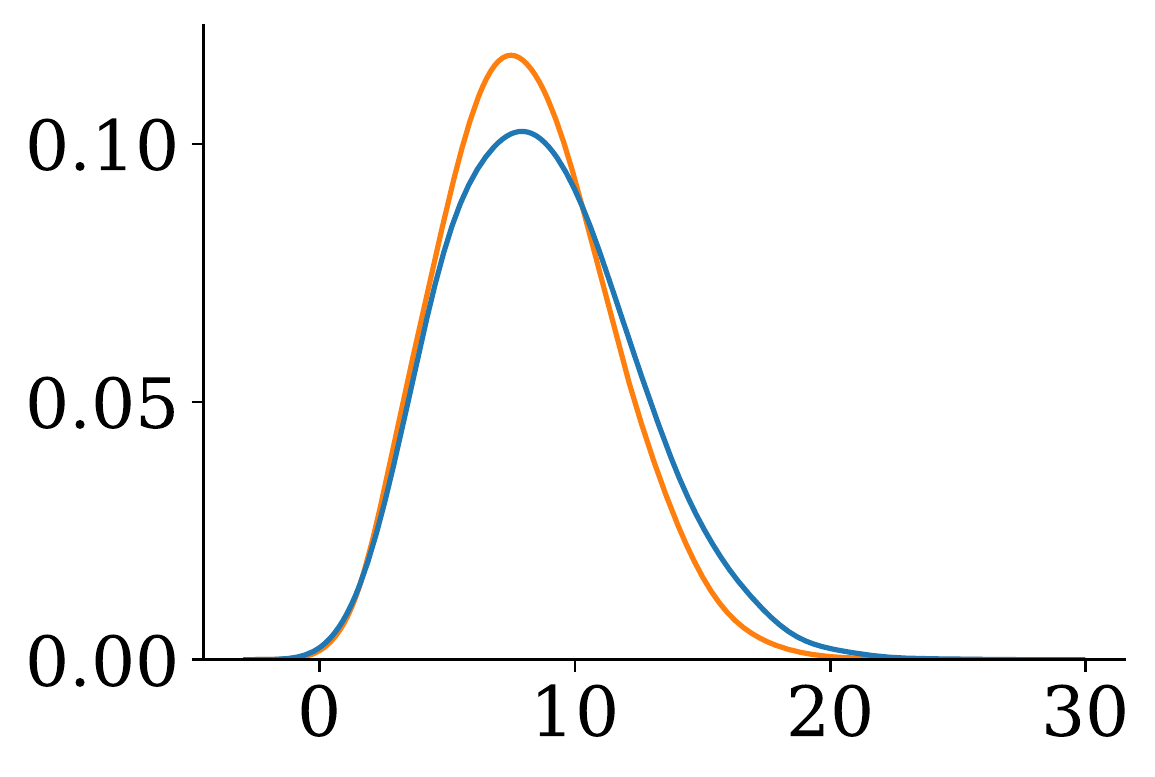}} &
\subfloat{\includegraphics[width=0.36\columnwidth, height = 3cm]{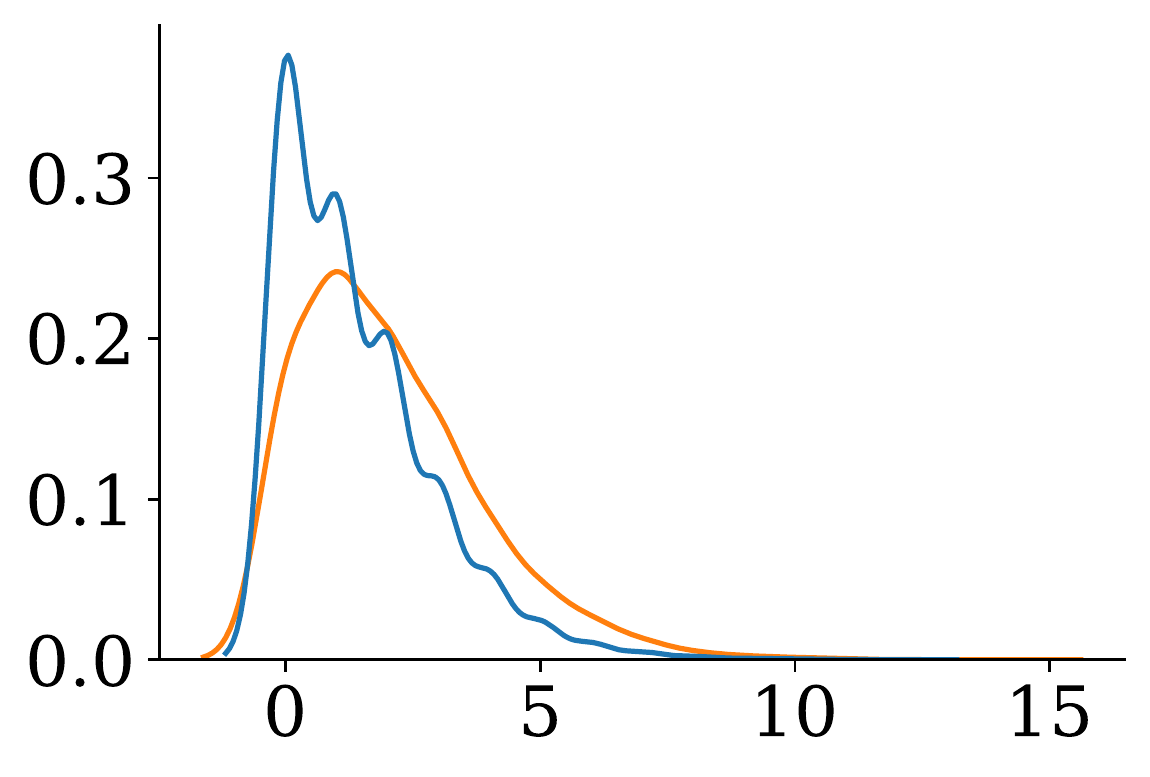}} &
\subfloat{\includegraphics[width=0.36\columnwidth, height = 3cm]{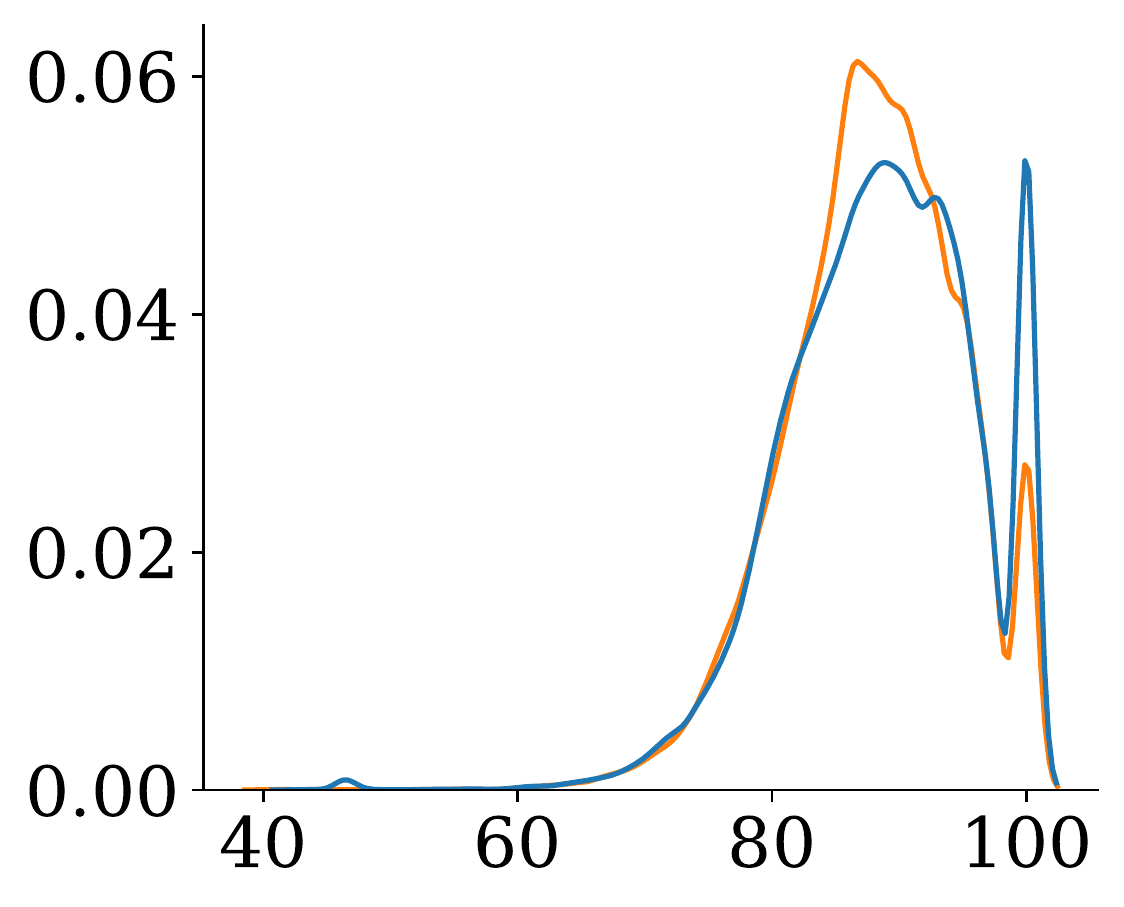}} &
\subfloat{\includegraphics[width=0.36\columnwidth, height = 3cm]{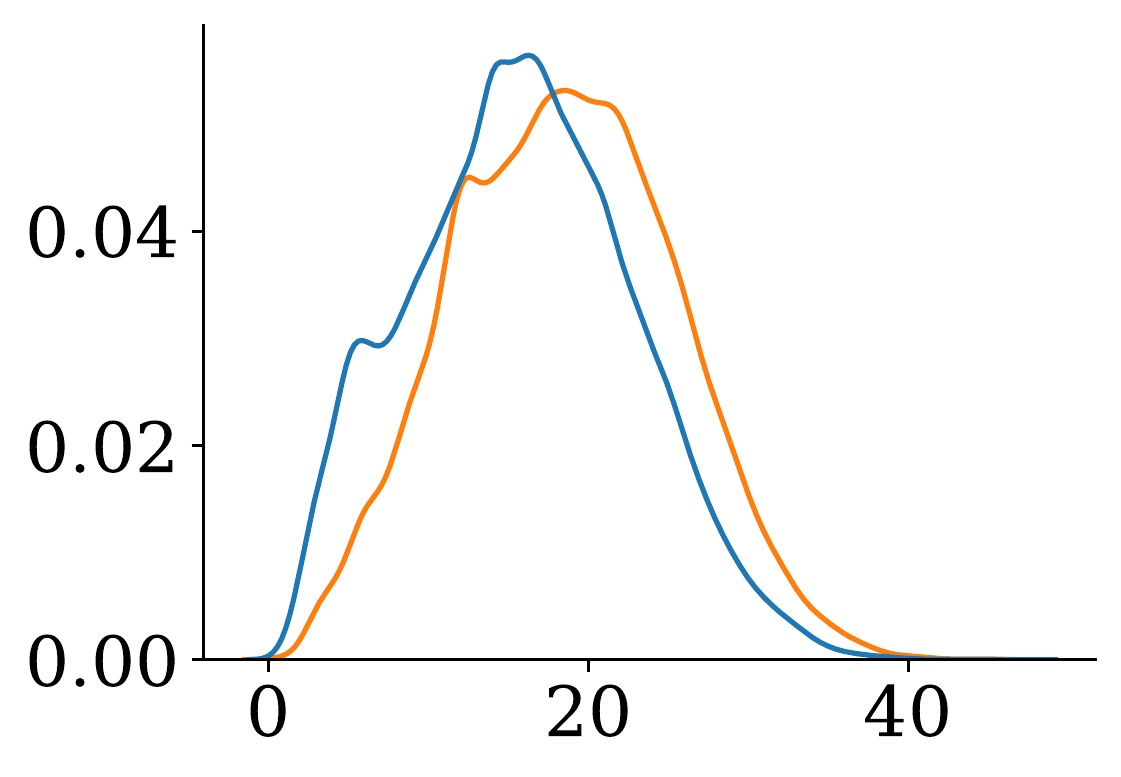}} &
\subfloat{\includegraphics[width=0.36\columnwidth, height = 3cm]{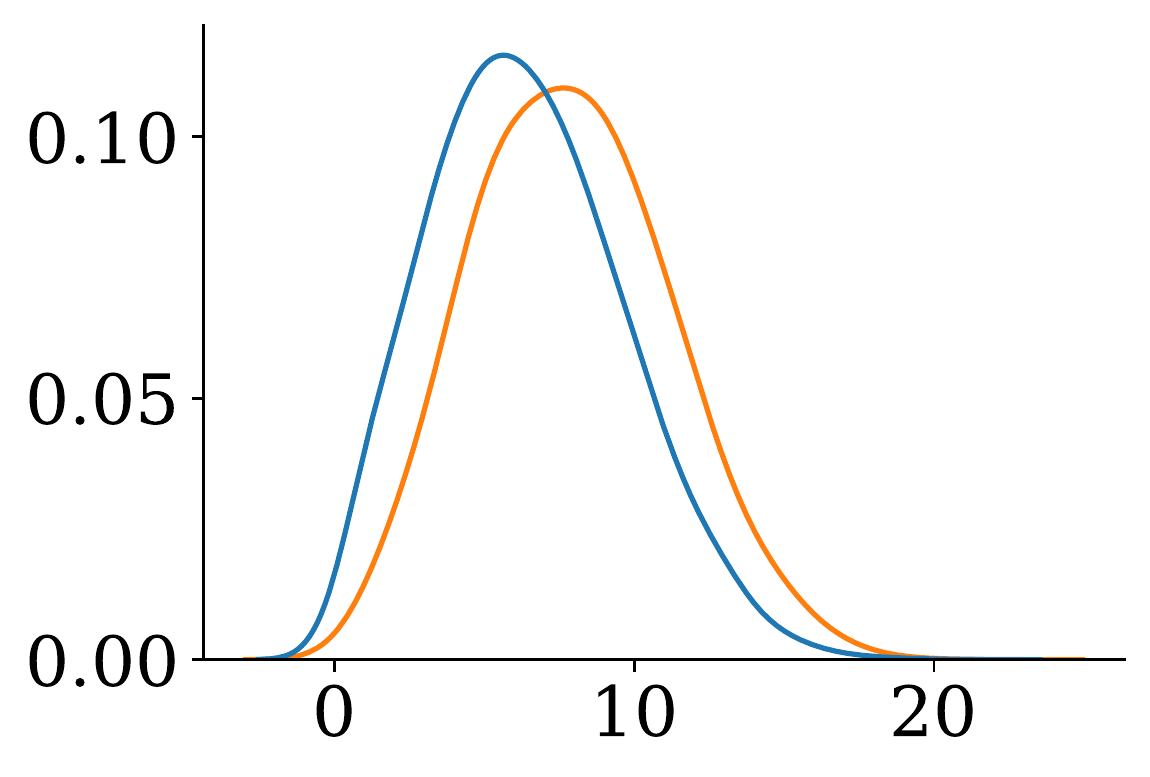}}\\\hline

(f) Conjunctions & (g) Adverbs & (h) Determiners & (i) Adjectives & (j) WH-words\\\hline
\subfloat{\includegraphics[width=0.36\columnwidth, height = 3cm]{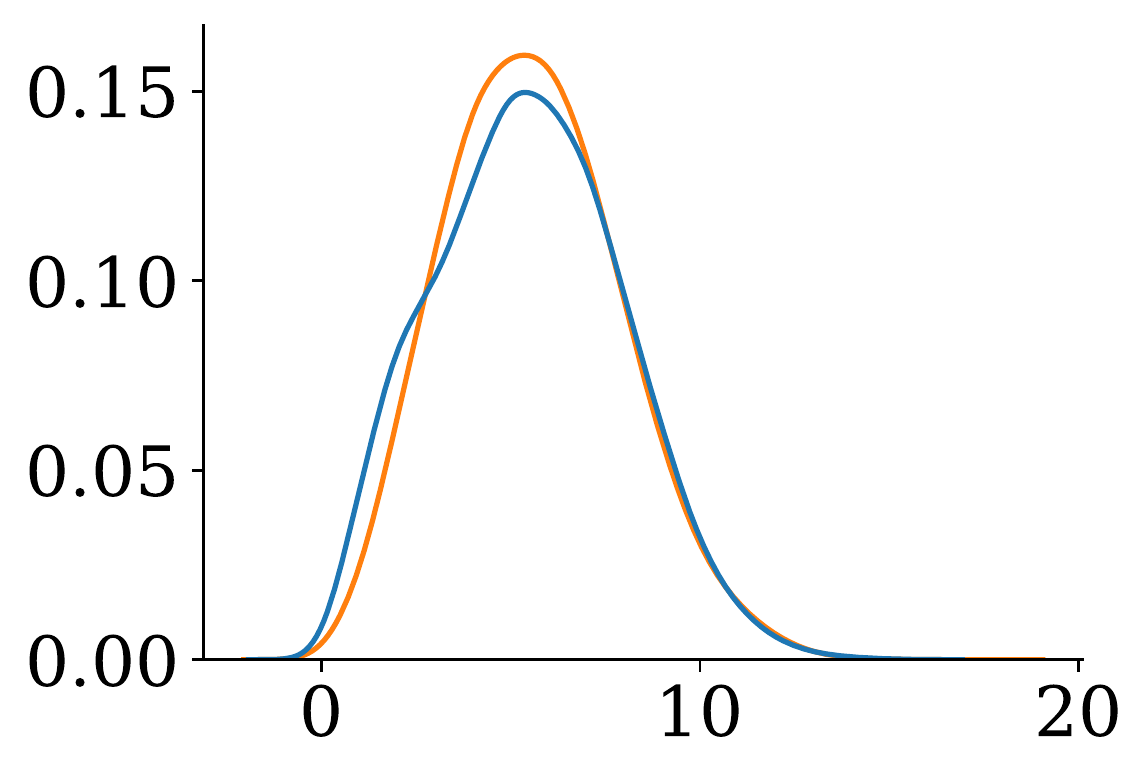}} &
\subfloat{\includegraphics[width=0.36\columnwidth, height = 3cm]{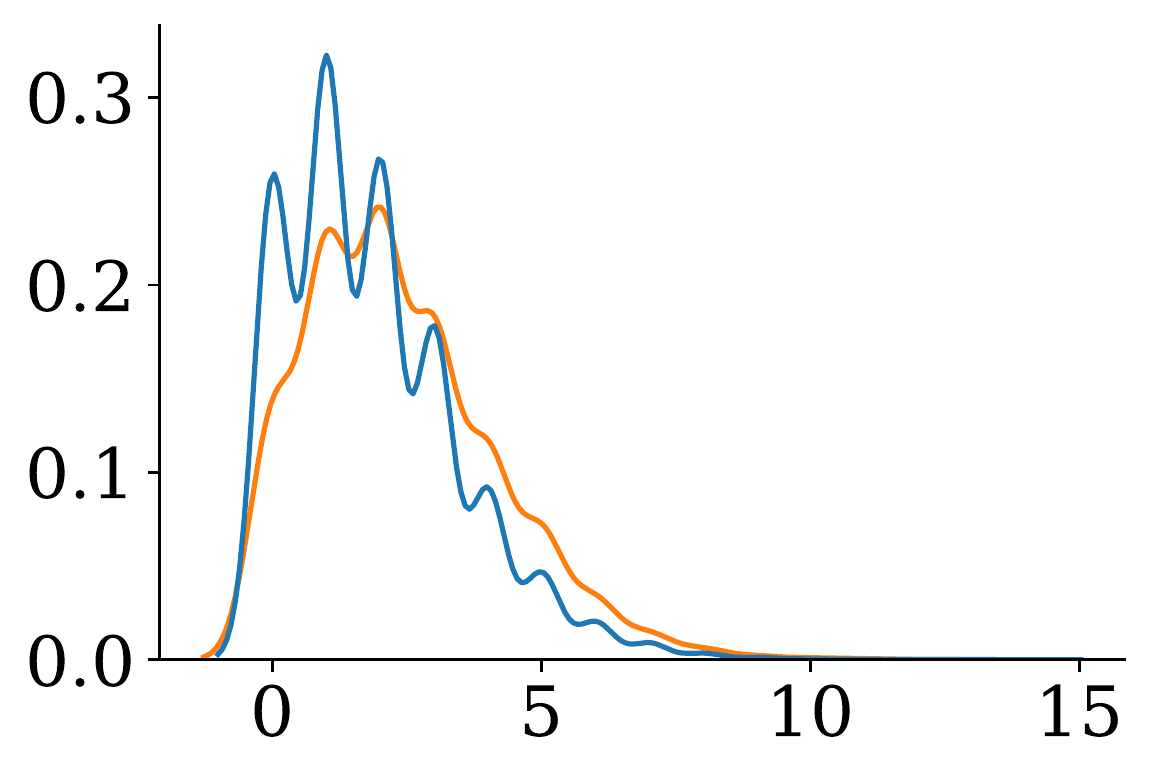}} &
\subfloat{\includegraphics[width=0.36\columnwidth, height = 3cm]{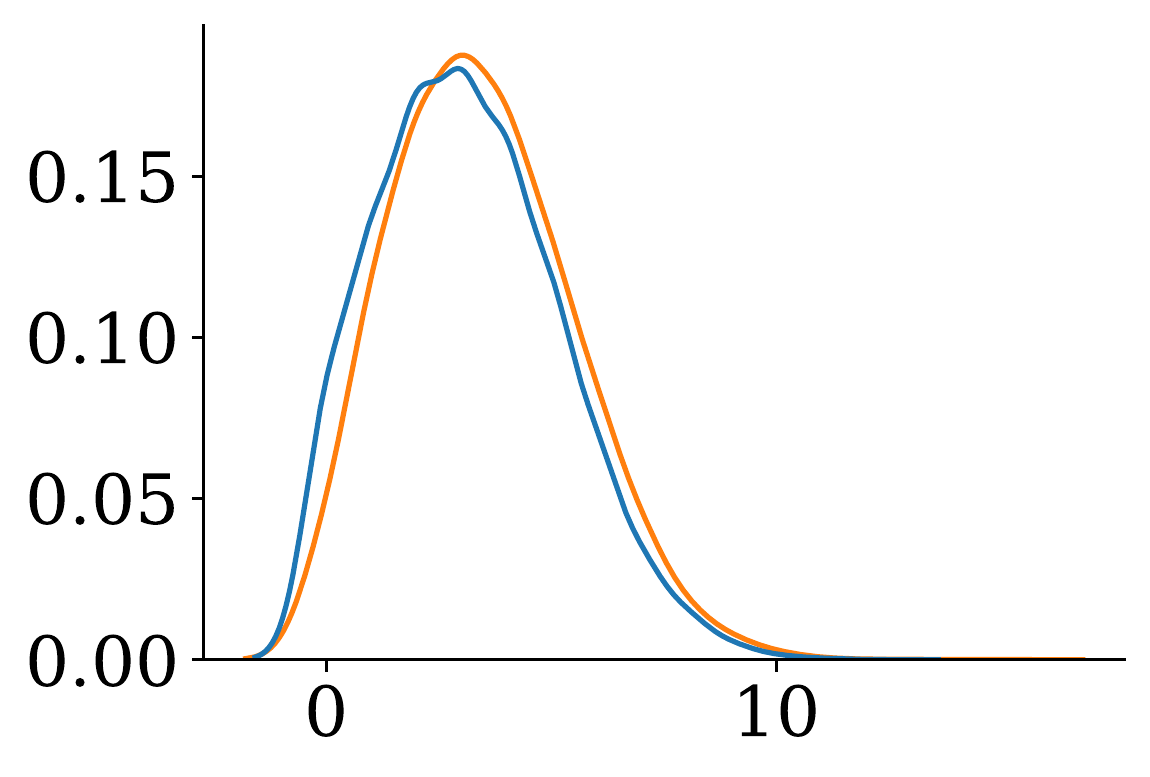}} &
\subfloat{\includegraphics[width=0.36\columnwidth, height = 3cm]{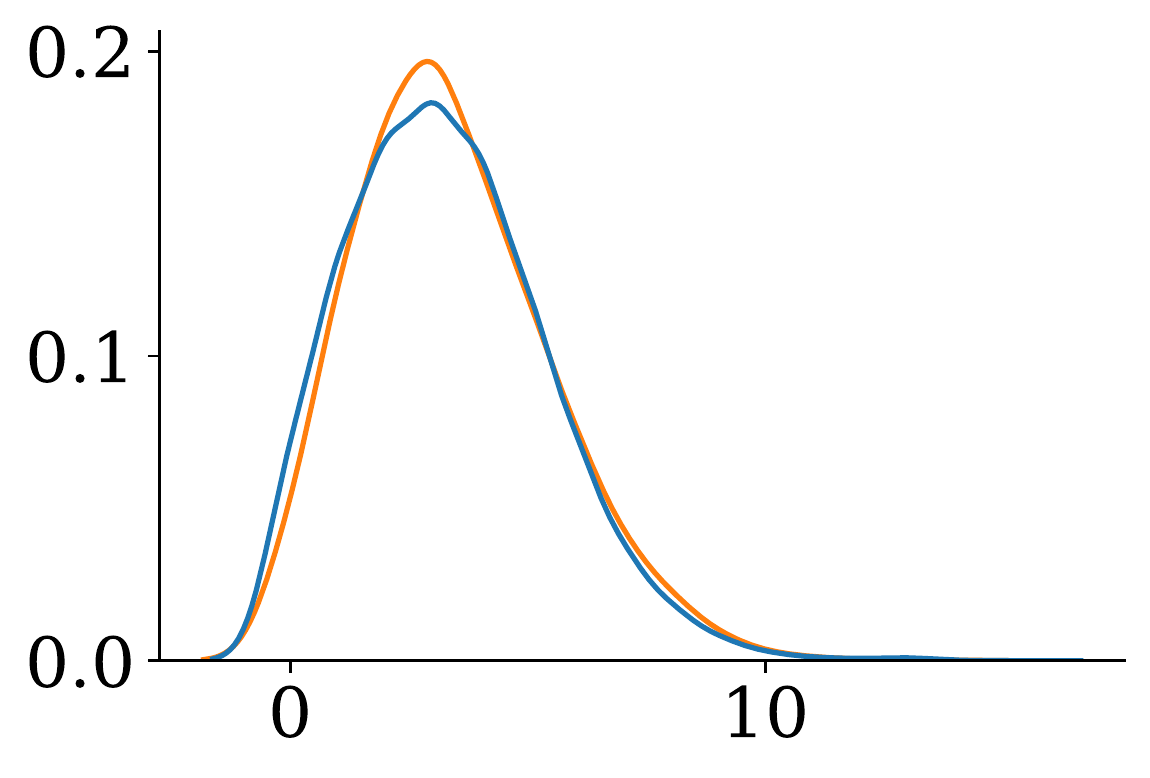}} &
\subfloat{\includegraphics[width=0.36\columnwidth, height = 3cm, trim={0cm 0cm 0cm 4cm}, clip]{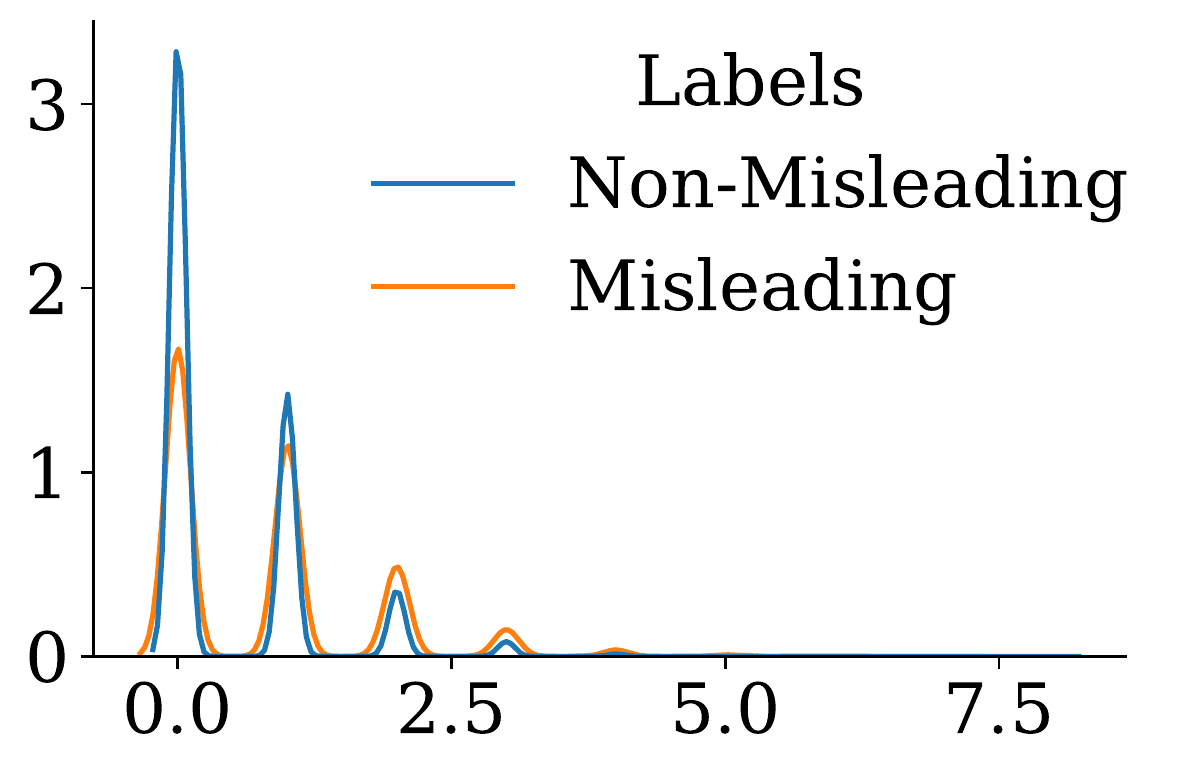}}\\\hline
\end{tabular}
\caption{Syntactic analysis \rev{of the Misleading and Non-Misleading tweets}. The x-axis represents the counts of the specific Syntactic attribute (for example, noun) and y-axis represents the density of the distribution. \rev{TTR is an abbreviation for Type-Token Ratio. \textbf{\textcolor{orange}{Orange}} and \textbf{\textcolor[HTML]{4682B4}{blue}} colors represent the Misleading and Non-Misleading labels, respectively. }}
\label{fig:linguistic}
\end{figure*}


First, we look at the \textbf{Nouns}, the main building blocks of any sentence.
We observe from Figure \ref{fig:linguistic}(a) that visually there is a slight variation in both distributions. To determine whether this difference in the distributions is statistically significant, we calculated the p-value of Nouns (see Table \ref{tbl:ks}, row 1), which is much lower than the significance level (5\% as default value), which implies that the two distributions are in fact dissimilar.
Second, \textbf{Pronouns} are the substitute for Nouns. Figure \ref{fig:linguistic}(b) shows that Pronouns are more used in \textit{Non-Misleading} tweets than \textit{Misleading} tweets.
Third, \textbf{Type-Token Ratio (TTR)} measures the lexical diversity (quality) of the text. Specifically, it is the ratio between the total number of unique words (types) in the text and the total number of words in the text. The higher the value of this ratio, the higher the lexical diversity of the text.
        \begin{equation}
            TTR = \frac{ \# \: of \: Types}{\# \: of \: Tokens} * 100
        \end{equation}
We notice in Figure \ref{fig:linguistic}(c) that the distributions are right-skewed, indicating that the text is of good quality in terms of lexical diversity. However, we observe some differences between the distributions. When TTR is near to 100, the density of the \textit{Misleading} distribution is lower in comparison to its mean. Whereas, \textit{Non-Misleading} distribution has a similar density with respect to its mean. This implies that \textit{Misleading} tweets are less lexically diverse in contrast to \textit{Non-Misleading} tweets. In addition, the p-value present in Table \ref{tbl:ks}, row 3 is lower, indicating the distributions are different.

    \begin{table}[!htbp]
    \caption{P-values of the Kolmogorov Smirnov Test \rev{for each of the syntactic characteristics extracted from the dataset.}}
    \label{tbl:ks}
    \centering
    \begin{tabular}{|c|c|}
    \hline
    \textbf{Syntactic Attributes} & \textbf{P-values}       \\ \hline
    \textbf{Nouns}                & 1.12e-109 \\ \hline
    \textbf{Pronouns}                  & 4.29e-78  \\ \hline
    \textbf{TTR}           & 2.09e-96  \\ \hline
    \textbf{Stop words}                & 1.10e-44  \\ \hline
    \textbf{Verbs}             & 7.81e-50   \\ \hline
    \textbf{Conjunctions}         & 5.89e-57   \\ \hline
    \textbf{Adverbs}              & 1.84e-25  \\ \hline
    \textbf{Determiners}          & 0.0                     \\ \hline
    \textbf{Adjectives}           & 0.0                     \\ \hline
    \textbf{WH-words}         & 0.0                     \\ \hline
    \end{tabular}%
    \end{table}

We investigate other Syntactic aspects namely, \textbf{Stop words} (Figure \ref{fig:linguistic}(d)), \textbf{Verbs} (Figure \ref{fig:linguistic}(e)), \textbf{Conjunctions} (Figure \ref{fig:linguistic}(f)), \textbf{Adverbs} (Figure \ref{fig:linguistic}(g)), \textbf{Determiners} (Figure \ref{fig:linguistic}(h)), \textbf{Adjectives} (Figure \ref{fig:linguistic}(i)), and \textbf{WH-words} (Figure \ref{fig:linguistic}(j)).
Their p-values present in the Table \ref{tbl:ks}.
Although the values for \textbf{Determiners}, \textbf{Adjectives} and \textbf{WH-words} are near but less than the significance level, we consider them distinguishable attributes in finding \textit{Misleading} and \textit{Non-Misleading} tweets.

\subsection{What are the Principal Topics of Discussion?}

Next, using topic modeling using the LDA approach \cite{xu2019detecting},
we inspect the top five most talked-about topics among \textit{Misleading} and \textit{Non-Misleading} tweets (see Table \ref{tbl:topics}).
We employed grid-search to identify the optimal number of topics, determining that to be five for both \textit{Misleading} and \textit{Non-Misleading} tweets for our data collection.




\noindent Following are the key themes of \textit{Misleading} tweets  -

\noindent \textit{1) Politics:}
The frequent targets of these tweets are politics; for example, the tweet - \textbf{``I don't even trust this Govt to take a vaccine they are desperate to sell us  This makes me feel sad But now if cellulitis is also a side effect of it And blood clots And Morrison brushes that aside and or lies to us about it What ?''}. Here, the key point of discussion is not to believe the government, and thus misleads the readers by bringing political angle into the vaccination debate and creating this type of vaccine prejudice. 


\noindent \textit{2) Myths and Side-Effects:}
The false stories and myths about vaccines have the greatest effect on people's minds; for instance, this tweet - \textbf{``IF you are allergic to eggs and chicken, you are not going to receive a dose of the 1.02 million doses of Oxford-Astrazeneca vaccine that is expected in Kenya on Tuesday 2 March, 2021.''} is attempting to steer readers away from the true story\footnote{https://www.unicef.org/kenya/press-releases/over-1-million-covid-19-vaccine-doses-arrive-nairobi-via-covax-facility}, namely the delivery of said volume of vaccines under the COVAX initiative.

\noindent \textit{3) Vaccine Efficacy}:
There is a lot of confusion related to the COVID-19 vaccine efficacy in the discussions, for instance, - \textbf{``How can we trust the vaccine when the efficacy is not reported precisely??''}.

\noindent \textit{4) Trump's Role:} 
Trump is well-known for being heavily involved in many false reports\footnote{https://www.cnbc.com/2021/01/13/trump-tweets-legacy-of-lies-misinformation-distrust.html}. This tweet confirms Trump's involvement in the tweet - \textbf{``Why did the whitehouse turn down Pfizer offer of the vaccine as the 1st to receive it? Seems Mr. Trump joins hands with Operation Warp Speed and they are deliberately slowing the speed of vaccine rollout.''} 

\noindent \textit{5) Vaccines Choice:}
There is skepticism about choosing the vaccines. For instance, this tweet tells one of the myths about altering DNA for the Johnson \& Johnson vaccine - \textbf{``Do not get the Johnson \& Johnson version.  The MOA is different than that of Pfizer's or Moderna's vaccine.  Picture a tennis ball inside a basketball.  J\&amp;J's enters the nucleus (tennis ball) that can permanently alter DNA.''}.
\begin{table}[!htbp]
\centering
\caption{Top 5  most discussed topics in Misleading and Non-Misleading COVID-19 vaccination tweets.}
\label{tbl:topics}
\begin{tabular}{|c|c|}
\hline
\textbf{Misleading}  & \textbf{Non-Misleading} \\ \hline
Politics     & Operation Warp Speed    \\ \hline
Myths \& Side-Effects      &  Shots          \\ \hline
 Vaccine Efficacy   &  Vaccine Efficacy      \\ \hline
Role of Trump        &  Real Side-Effects         \\ \hline
Vaccine Choices        &  Data \& Facts           \\ \hline
\end{tabular}%
\end{table}
\noindent The key themes of \textit{Non-Misleading} tweets are:

\noindent \textit{1) Operation Warp Speed:} The most discussed topic - Operation Warp Speed, initiated by the US government to facilitate the development and distribution of COVID vaccines. An example of a tweet with this theme - \textbf{``[username] Pfizer vaccine was self funded. Nothing to do with warp speed, the government or the lying ex president trump.  TRUMP IS A LIAR''}. 

\noindent \textit{2) Shots \& Real Side-Effects:}
These two themes are about informing an individual's actual experience after getting the vaccine shot. We provide an example of a tweet that is related to both the themes - \textbf{``I had the PFIZER shot(s), 30 days apart, and both times it didn't hurt. My arm was sore later on for a day or two. 24 hours after the second shot I got a fever and chills and promptly went to sleep. Woke up 10 hours later feeling great. No other side effects. GET THIS VACCINE!''}

\noindent \textit{3) Vaccine Efficacy:}
Both \textit{Misleading} and \textit{Non-Misleading} have a shared theme. This indicates that they are both discussing the vaccine's efficacy. \textit{Misleading} tweets, on the other hand, aim to cause uncertainty regarding vaccine efficacy, while \textit{Non-Misleading} tweets emphasize the positive aspects of vaccine efficacy, such as - \textbf{``\#JabMe: A single shot of either the Pfizer or Oxford vaccine provides about 80 percent protection against being treated in a hospital, according to the latest data from the UK vaccination program.''}

\noindent \textit{4) Data \& Facts:} 
\textit{Non-Misleading} tweets are more concerned with presenting accurate information, such as - \textbf{``BBC: Around 5 million Europeans have already received the AstraZeneca vaccine.  Of this figure, about 30 cases had reported "thromboembolic events" - or developing blood clots.  European medicines regulator said there was no indication the jab was causing the blood clots.''}

The top five themes indicate the different subjects explored in both types of tweets. Precisely, \textit{Misleading} tweets mostly mislead the reader using political dimension or raise fear among the people for vaccination. In comparison, \textit{Non-Misleading} tweets discuss the real side-effects of the vaccination and try to bring the facts with evidence.

\section{Opinion Study}\label{sec:persona}
Previously, we explored the Syntactic dimension. We now look into the second dimension, the role of opinion in relation to \textit{Misleading} and \textit{Non-Misleading} tweets.

\subsection{Sentiment Matters}
We first explore the impact of sentiments on \textit{Misleading} and \textit{Non-Misleading} tweets, considering three broad categories of sentiments: Positive, Negative, and Neutral. The sentiments are calculated using VADER API \cite{yaqub2020tweeting,waters2021exploring}. 
To detect a tweet as Positive, Negative, or Neutral, we utilized the compound score. The compound score is the sum of the valence scores of each word, yielding a range of [-1, 1]. A score of -1 indicates a strong Negative sentiment, whereas a score of +1 suggests a strong Positive sentiment. We utilized a 0.05 threshold value. 
\rev{
The selection of the 0.05 threshold has been adopted from the previous literature \cite{bonta2019comprehensive, cruz2021analyzing}. Additionally, to validate the choice of threshold, we randomly sampled 150 tweets with a balanced combination of Positive, Negative, and Neutral sentiment labels obtained from the VADER API (50 random tweets from each label). Following that, we manually annotated the sentiments for 150 tweets and got an accuracy of 0.96, confirming the efficacy of the selected threshold.
}
Thus, a tweet is considered to have a Positive sentiment if the compound score is greater than or equal to 0.05, neutral if the score is between 0.05 and -0.05; Negative otherwise.


\begin{figure}[!htbp]
    \centering
    \includegraphics[width=8.7cm ]{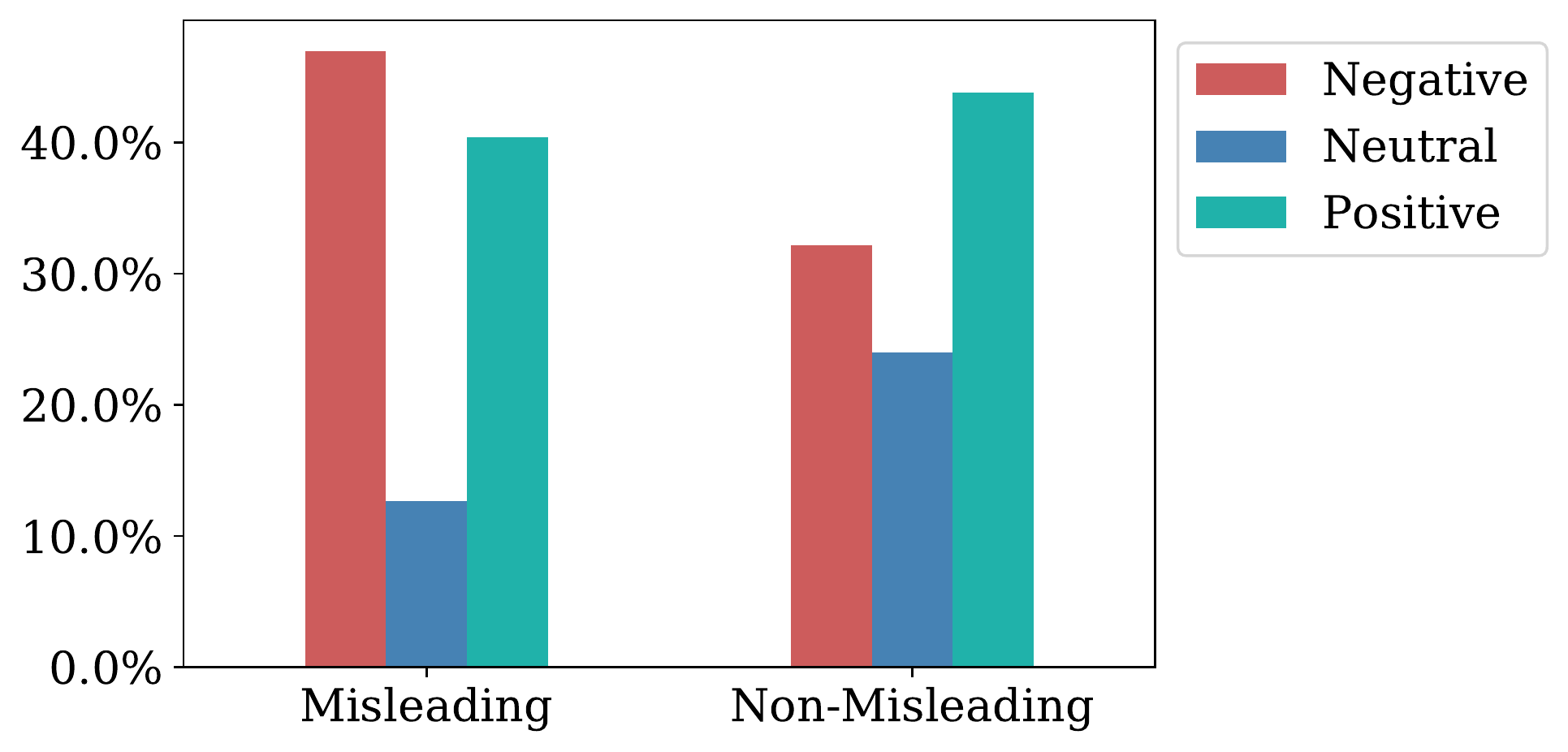}
    \caption{Sentiment analysis \rev{with respect to Misleading and Non-Misleading tweets.} The x-axis and y-axis denote the labels of the tweets and percentages, respectively. \rev{\textbf{\textcolor[HTML]{E34234}{Negative}}, \textbf{\textcolor[HTML]{4682B4}{Neutral}}, and \textbf{\textcolor[HTML]{088F8F}{Positive}} are different categories of the sentiments.}}\label{fig:senti}
\end{figure}

Figure \ref{fig:senti} shows that Negative sentiments are more prevalent in \textit{Misleading} tweets followed by Positive and Neutral sentiments. 
An example of a \textit{Misleading} tweet with Positive sentiment is \textbf{``A little Angel in my dreams today told that our bodies will be developing antibodies on its own within a few days without vaccination''}. While a reading of it indicates vaccine prejudice and skepticism, the sentiment analysis tool latches upon the positive sounding phrases in there. 

We then identify the topics which are being discussed with respect to sentiments. Apart from Vaccine Efficacy which confirms the above mentioned tweet's theme, the related topics Operation Warp Speed and Trials are also identified within the Positive sentiments of \textit{Misleading} tweets (See Table \ref{tbl:topic-sent}).
We surmise that \textit{Misleading} tweets with Positive sentiments inject the negativity by sugar-coating the tweets with positive words to easily trick people into either believing in their positive hypothetical situations or providing a new dimension to the topic.


Positive sentiments, on the other hand dominate in \textit{Non-Misleading} tweets, though a substantial portion of them again have Negative or Neutral sentiments. An instance of a \textit{Non-Misleading} tweet with Negative sentiment is \textbf{``Dr Kathrin Jansen, Pfizer's head of vaccine development: We were never part of the Warp Speed ... We have never taken any money from the U.S. government, or from anyone. Trump is a liar''}. 
In this instance, the Negative sentiment of the \textit{Non-Misleading} tweet is due to it counteracting the \textit{Misleading} information. Many facts and news related to the pandemic have naturally Negative sentiments.
Similar to this tweet argument, topics that are discovered in the Negative sentiments of \textit{Non-Misleading} tweets are Operation Warp Speed and Vaccine Efficacy, in addition to, Trials and Data \& Facts (See Table \ref{tbl:topic-sent}). 
These \textit{Non-Misleading} tweets with Negative sentiments indicate that they are either attempting to clarify claims against COVID Vaccination's Development Companies or myths against the vaccination process with their choice of negative words.


\begin{table}[!htbp]
\centering
\caption{Topic Modeling with respect to each Sentiments. M and NM represent Misleading and Non-Misleading. OWS and VaEf denote Operation Warp Speed and Vaccine Efficacy, respectively.}
\label{tbl:topic-sent}
\begin{tabular}{l|l|l|l|}
\cline{2-4}
                         & Positive                                                                        & Negative                                                                       & Neutral                                                                              \\ \hline
\multicolumn{1}{|l|}{M}  & \begin{tabular}[c]{@{}l@{}}Trials, OWS, \\ VaEf\end{tabular}                    & \begin{tabular}[c]{@{}l@{}}VaEf, OWS, \\ Trials, Myths\end{tabular}            & \begin{tabular}[c]{@{}l@{}}VaEf, OWS, \\ Trials\end{tabular}                         \\ \hline
\multicolumn{1}{|l|}{NM} & \begin{tabular}[c]{@{}l@{}}Trump, \\ Real \\ side-effects, \\ VaEf\end{tabular} & \begin{tabular}[c]{@{}l@{}}Data \& Facts, \\ OWS, \\ VaEf, Trials\end{tabular} & \begin{tabular}[c]{@{}l@{}}Data \& Facts, \\ VaEf, \\ Real side-effects\end{tabular} \\ \hline
\end{tabular}
\end{table}
Overall, Negative sentiments are more common in \textit{Misleading} tweets, whereas, \textit{Non-Misleading} tweets have more Positive sentiments. 
We go through five emotions in detail in the next Section.

\subsection{Intense Emotions}
The sentiments serve as the foundation for analyzing the tweets. As a result, we dig deeper into the impact of emotions on tweets.

\begin{table*}[!htbp]
\centering
\caption{Topic modeling with respect to each five emotions. M and NM represent Misleading and Non-Misleading, respectively. \rev{OWS and VaEf are abbreviations for Operation Warp Speed and Vaccine Efficacy, respectively.}}
\label{tbl:topic-emo}
\begin{tabular}{l|l|l|l|l|l|}
\cline{2-6}
                         & Fear                                                             & Surprise                                                                   & Sadness                                                                  & Anger                                                                                              & Happiness               \\ \hline
\multicolumn{1}{|l|}{M}  & \begin{tabular}[c]{@{}l@{}}Trials, OWS, VaEf\end{tabular}     & \begin{tabular}[c]{@{}l@{}}Trials, Politics, Trump, Myths\end{tabular}  & \begin{tabular}[c]{@{}l@{}} VaEf, OWS, Politics \end{tabular}                                                      & \begin{tabular}[c]{@{}l@{}}VaEf, OWS, \\ Availability, Trials\end{tabular}                         & \begin{tabular}[c]{@{}l@{}} VaEf, approval \end{tabular}         \\ \hline
\multicolumn{1}{|l|}{NM} & \begin{tabular}[c]{@{}l@{}}Trump, \\ Politics, VaEf\end{tabular} & \begin{tabular}[c]{@{}l@{}} Shots, Data \& Facts, OWS, VaEf\end{tabular} & \begin{tabular}[c]{@{}l@{}}Data \& Facts, VaEf, Politics\end{tabular} & \begin{tabular}[c]{@{}l@{}}VaEf, Data\\ \& Facts, \\ Real side-effects, \\ Availability\end{tabular} & \begin{tabular}[c]{@{}l@{}} VaEf, Real side-effects \end{tabular} \\ \hline
\end{tabular}
\end{table*}

\begin{figure}[!htbp]
    \centering
    \includegraphics[width=10cm ]{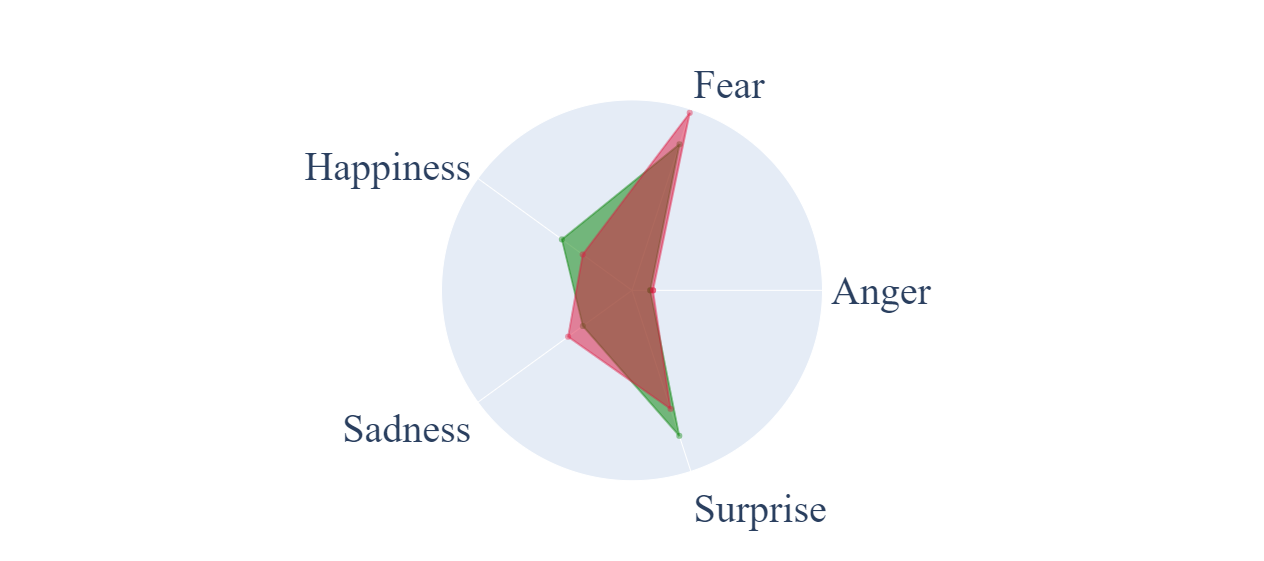}
    \caption{Emotion analysis \rev{of the Misleading and Non-Misleading tweets. Each axis denotes the emotion. The five emotions depicted are Happiness, Fear, Anger, Surprise, and Sadness.} Non-Misleading tweets are represented in \textbf{\textcolor[rgb]{0.1, 0.6, 0.3}{green}} color, and Misleading in \textbf{\textcolor[rgb]{0.8, 0.0, 0.0}{red}} color (best seen in color).}\label{fig:emotions_radar}
\end{figure}

Figure \ref{fig:emotions_radar} displays the five different emotions - Anger, Fear, Happiness, Sadness, and Surprise retrieved by employing the text2emotion API \cite{dhar2020emotions}.
In this API, for each tweet, we get a dictionary where keys denote emotion categories, and values indicate its score. Higher the score, the higher the emotion in that tweet. We classify the tweets as per the emotion that has the highest score. In the Figure, the value with respect to each emotion axis corresponds to the percentage of a particular emotion in a \textit{(Non-)Misleading} class.


In \textit{Misleading} tweets, the most common emotion is Fear, followed by Surprise, Sadness, Happiness, and, finally, Anger. Whereas, \textit{Non-Misleading} tweets have a tie for the first place with Fear and Surprise, followed by Happiness, Sadness, and Anger.

Fear and Surprise are the two most popular emotions across both types of tweets. It is understandable given that 45\% of unvaccinated people are afraid to get the vaccine because they are worried about the adverse side-effects\footnote{https://www.vox.com/recode/22330018/covid-vaccine-hesitancy-misinformation-carnegie-mellon-facebook-survey}. 
To confirm this, we studied the topics around Fear and Surprise emotions. The topics which are similar to the above statement are Trials and Vaccine Efficacy in both \textit{Misleading} and \textit{Non-Misleading} categories (See Table \ref{tbl:topic-emo}).
However, the emotion Fear is higher in \textit{Misleading} tweets in contrast to \textit{Non-Misleading} tweets, which is attributable to the observation that most \textit{Misleading} tweets reference fake and fabricated vaccine side-effects and false narratives of Operation Warp Speed (Table \ref{tbl:topic-emo}).
In contrast, the emotion Surprise is higher in \textit{Non-Misleading} tweets than \textit{Misleading} tweets. These often discuss the governments' fast response towards vaccination, 
fitting into the Data \& Facts topic.

Furthermore, emotions, Anger, and Sadness are higher in \textit{Misleading} tweets. One of the possible reasons could be that these tweets often involve a political dimension and accuse the government of not making the right decisions.
The topics covering under both emotions are Politics, Vaccine Availability, and Operation Warp Speed in the \textit{Misleading} category.
\textit{Non-Misleading} tweets that have emotion Happiness discuss their experience about receiving the shot and facing no bogus side-effects spreading across the Internet. 
A related topic is Real side-effects in the \textit{Non-Misleading} category.
To summarize, the majority of the \textit{Misleading} tweets have Fear emotions more than \textit{Non-Misleading} tweets.

\section{The Influence of Visibility}\label{sec:visibility}
So far, our analysis was confined to the content of the tweets themselves. The focus of the third dimension is looking at information and meta-data in the tweets that influence their visibility, such as words used, hashtags, likes to study whether there are distinctive characteristics across \textit{Misleading} and \textit{Non-Misleading} tweets.

\subsection{The merry words of Twitter}
\begin{figure}[!htbp]
        \centering
       \subfloat[Non-Misleading Tweets  \label{subfig-2:wc_n}]{%
       \includegraphics[width=0.26\textwidth]{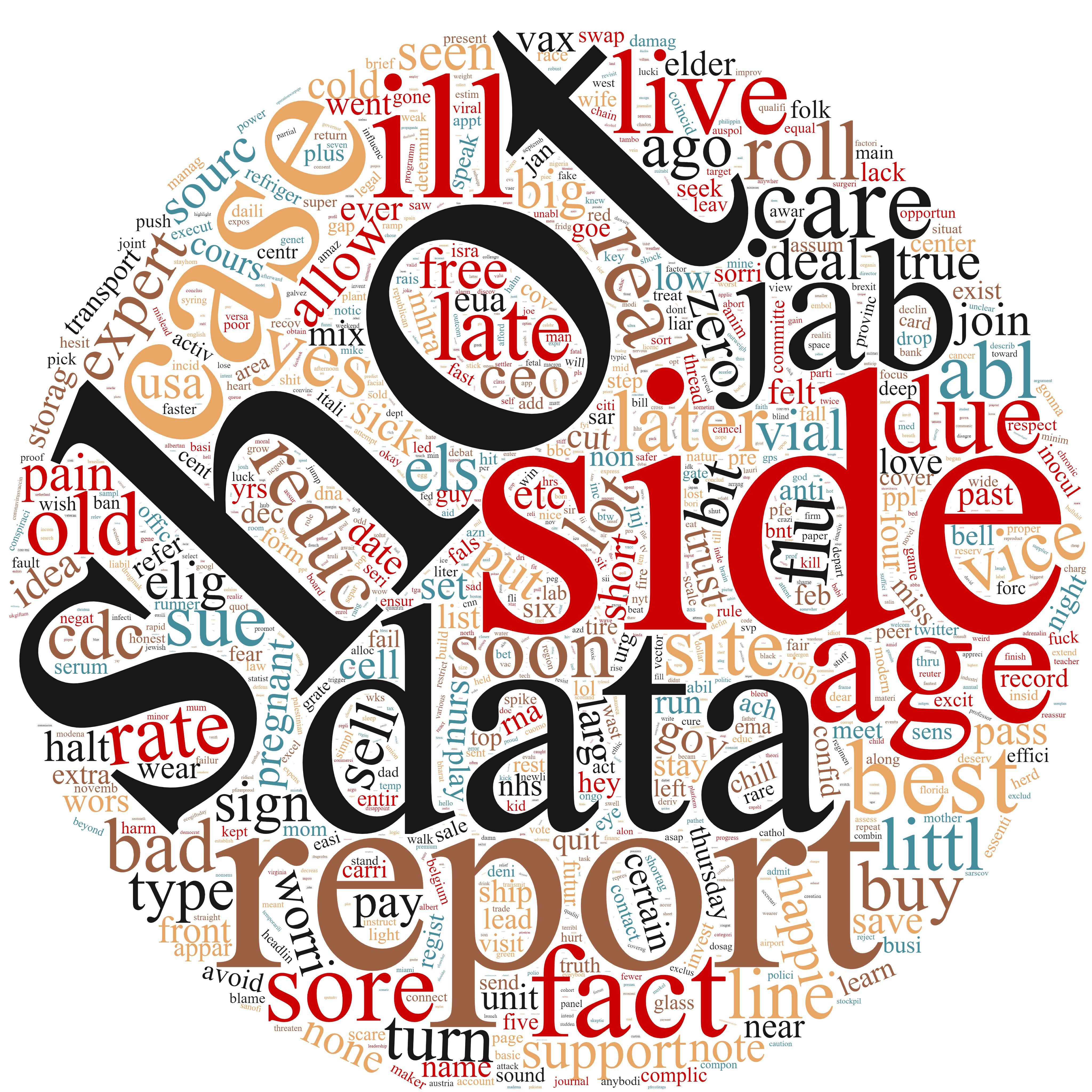}
     }
     \subfloat[Misleading Tweets \label{subfig-1:wc_m}]{%
       \includegraphics[width=0.26\textwidth]{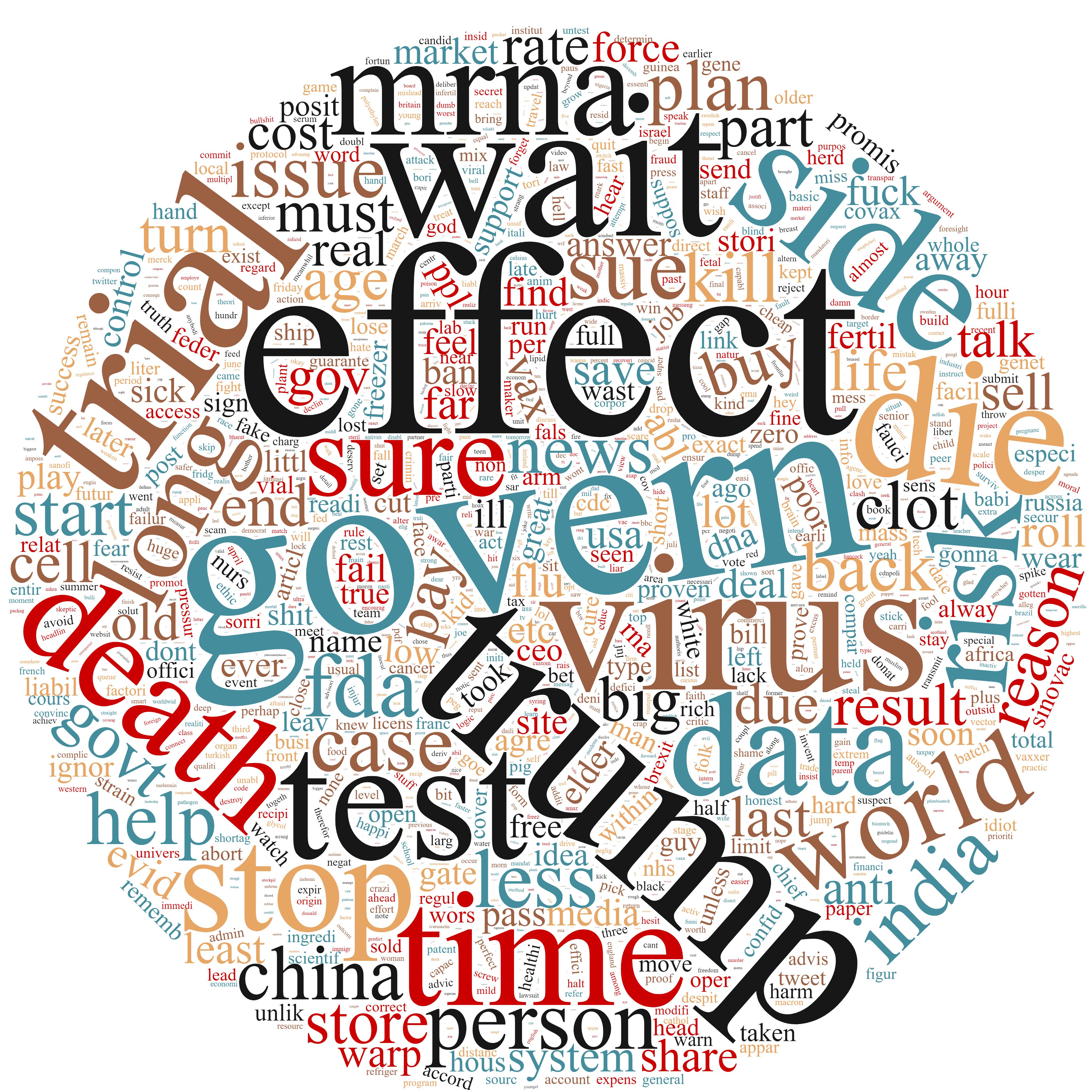}
     }
     \caption{Word Clouds \rev{with respect to Non-Misleading and Misleading tweets.} The size of the word is proportional to its frequency.}
     \label{fig:aucroc}
   \end{figure}

Certain words are used more frequently in the tweets than others. In Figures \ref{subfig-2:wc_n} and \ref{subfig-1:wc_m}, we use Word Clouds to summarize this for both types of tweets visually. The relative frequency of the words is reflected in the size of the words.
The Figures show that the most frequent words in both Word Clouds are completely different, indicating that the choice of the words in both types of tweets significantly varies. \textit{Shot, report, jab, ill, sore,} and \textit{fact} are all recurring words in the \textit{Non-Misleading} Word Cloud, whereas, \textit{wait, death, risk, effect, trump,} and \textit{die} are all frequent words in the \textit{Misleading} one.

To study this difference quantitatively, we computed the Kendall Tau correlation coefficient\footnote{https://online.stat.psu.edu/stat509/lesson/18/18.3} on the union of the top 50 \textit{Misleading} and \textit{Non-Misleading} words.
Only seven words were found to be common in both classes. 
A score of -0.81 was observed, showing disagreement between the word groups (the Kendall Tau range is [-1, 1], with -1/1 indicating strong dis/agreement respectively). This clearly suggests that frequently recurring words are unrelated, implying that the word choices in both classes differ.

\subsection{Much ado about Hashtags}

Extensive usage of hashtags is a popular way to enhance the exposure of the tweets. We investigate them from two perspectives.

\noindent \textbf{Unique Hashtags:} We explore such hashtags that are relatively unique to \textit{Misleading} versus \textit{Non-Misleading} tweets. Table \ref{tbl:hashtags} lists some of the popular ones. Note that the hashtags mentioned in the Table are chosen depending on how many times they appear in the tweets. In \textit{Misleading} tweets, the \#untestedvaccine clearly indicates that the tweet refers to one of the vaccine myths. In contrast, the \#vaccinatedandproud represents that tweet is in support of the vaccination process. Thus, the choice of the hashtags can provide a clue about the \textit{Misleading} tweets.
    
\noindent \textbf{Co-hashtags:} We also consider the combination of hashtags that frequently occurred together in a tweet, i.e., co-hashtags. In \textit{Non-Misleading} tweets, we find 280 co-hashtags, while 86 co-hashtags are found in \textit{Misleading} tweets. After filtering those co-hashtags that occurred more than once, we found that co-hashtags repeatedly occurred only in \textit{Non-Misleading} tweets. There is no pattern (consistency) concerning co-hashtags in \textit{Misleading} tweets, making their hashtags more random.

    \begin{table}[!htbp]
    \centering
    \caption{Use of 10 different Hashtags (left-side represents the hashtags that are mostly present in Non-Misleading tweets but not in Misleading tweets, and vice-versa on the right).}
    \label{tbl:hashtags}
    \begin{tabular}{|c|c|}
    \hline
    \textbf{Non-Misleading}     & \textbf{Misleading} \\ \hline
   fullyvaccinated    &   saynotopoisonvaccines       \\ \hline 
    savetheplanet         &      vaccineextortion      \\ \hline
        healthnews       &   pseudoscience            \\ \hline
    thisismyshot &       trumpvirusdeathtoll240k      \\ \hline
     covid19updates        &      untestedvaccine     \\ \hline
     scienceisreal  &   iwillnotgetvaccinated         \\ \hline
       publichealth      &    billgatesisevil         \\ \hline
       inthenews     &     abolishbigpharma           \\ \hline
         vaccinatedandproud       &  novaccine4me     \\ \hline
        2ndshot   &       astrazenecapoison           \\ \hline
    \end{tabular}%
    \end{table}

\subsection{As you Like it}
\begin{figure}[!htbp]
    \centering
    \includegraphics[width=6cm, height = 4cm, trim = {0, 0, 0, 0cm} ]{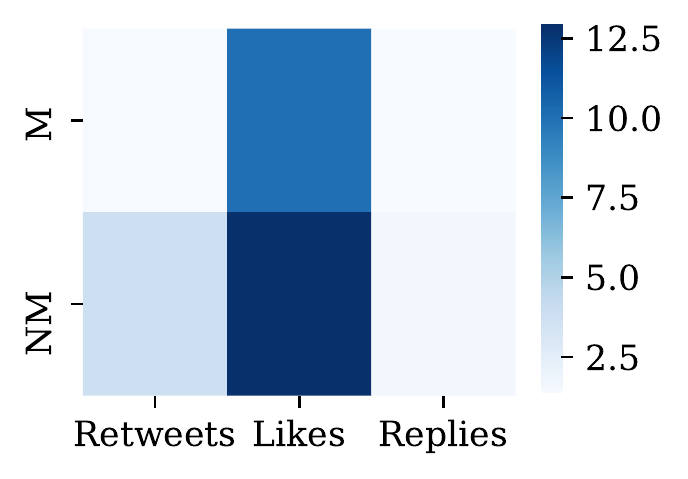}
    \caption{Aggregation of Visibility Counts. Each cell represents the median values of the Retweets, Replies, and Likes with respect to the labels of tweets. \rev{NM and M denote the Non-Misleading and Misleading tweets, respectively.}}\label{fig:heatmap_retweet}
\end{figure}

The number of Retweets, Replies, and Likes count are all essential visibility attributes. Figure \ref{fig:heatmap_retweet} depicts the median (mean) values of the counts of Retweets, Replies, and Likes for both classes. In our case, the median and mean values are the same.
On average, Replies to the tweets remain the same regardless of the type of information it contains. However, there is a variation in the Retweets count and Likes count. Relatively, the \textit{Misleading} tweets get fewer Retweets and Likes than \textit{Non-Misleading} tweets.


\subsection{What's in a Name?}

The names of the COVID-19 vaccinations are frequently mentioned in tweets. In this regard, we attempt to assess the influence of vaccine names in both \textit{Misleading} and \textit{Non-Misleading} tweets. 
In our collected data, we discovered that the majority of the discussions revolve around the five vaccinations: Pfizer, Moderna, AstraZeneca, Covaxin, and Johnson $\&$ Johnson. These names are used either in an individual or combined manner.
\begin{figure}[!htbp]
\centering
       \subfloat[]{%
       \includegraphics[width=0.26\textwidth, height = 3cm]{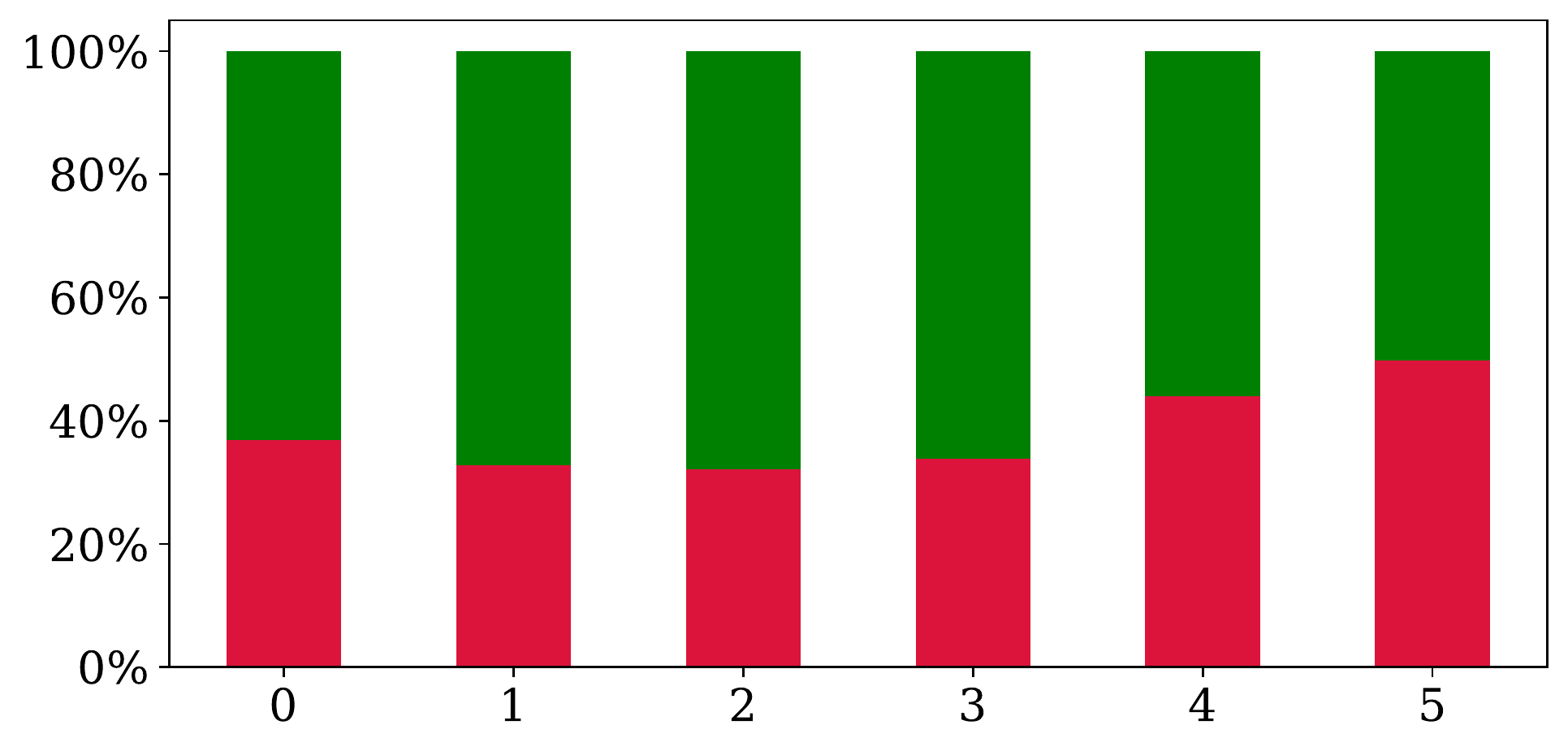}
     }
     \subfloat[]{%
       \includegraphics[width=0.26\textwidth]{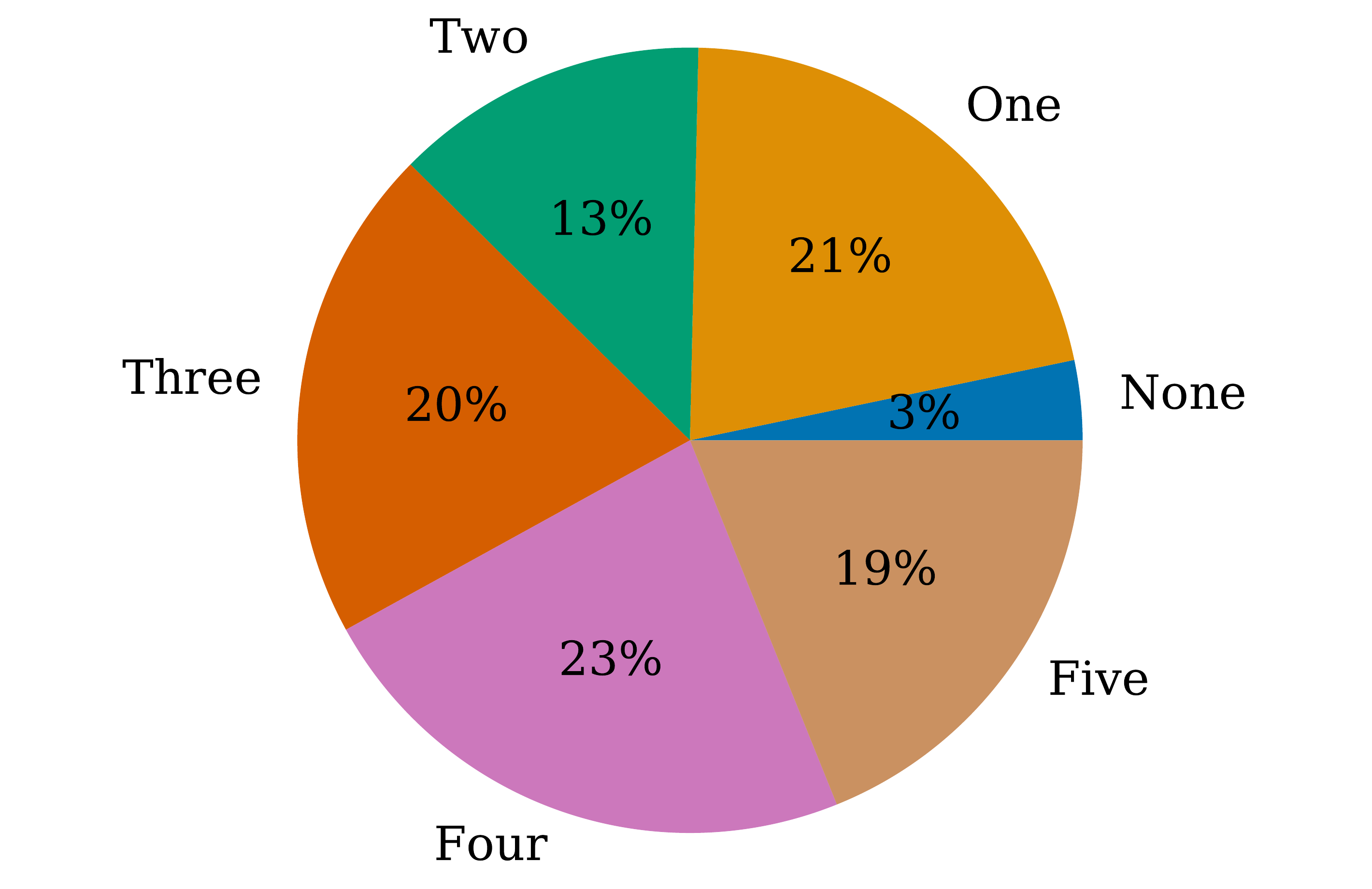}
     }
     \caption{Count of Vaccine Names used in a Tweet. X-axis and y-axis in (a) represents the count of the vaccines' names and percentages, respectively. The \textcolor[rgb]{0, 0.5, 0.1}{green} and \textbf{\textcolor[rgb]{0.8, 0.0, 0.0}{red}} colors indicate the Non-misleading and Misleading classes.  
     Value 0 on the x-axis corresponds to no mention of the vaccine name in the tweet, Value 1 denotes mention of one vaccine, and so on. (b) shows the percentage of the tweets with respect to the count of the vaccine name (best seen in color). 
     }
     \label{fig:vaccines_names_overlap}
   \end{figure}

Figure \ref{fig:vaccines_names_overlap} shows that the proportion of \textit{Misleading} tweets is lower than the proportion of \textit{Non-Misleading} tweets until the number of vaccine names is fewer than or equal to three. When the count reaches four or five, the number of \textit{Misleading} tweets begins to rise, i.e., when the number of vaccine names in a tweet grows to more than three, the likelihood of a \textit{Misleading} tweet also grows.

\section{Classification of Misleading Tweets}\label{sec:Exp}


By providing the tweets as an input to the pre-trained XLNet model, the focus was to obtain the labels for the tweets. These labels are treated as `ground-truth' for the prediction task.
In this section, instead of using the tweets themselves, we use the descriptive features described in the previous sections as an input to the various machine learning models to classify \textit{Misleading} and \textit{Non-Misleading} tweets. The purpose is to check if these features can distinguish \textit{Misleading} tweets. This helps us explicitly understand the differentiating characteristics across Non-Misleading tweets. 
Specifically, the features constitute - \textit{Stop words, Pronouns, Nouns, Adjectives, Average length, WH-word, Adverbs, Conjunctions, Verbs, Determiners, TTR, Sentiments, Emotions, and Hashtags}. 
For classification, we treat distinct sentiment categories as one feature by utilizing their raw scores. In emotions, there are dictionaries where keys denote emotion categories, and values indicate their scores. Higher the score, the higher the emotion in that tweet. Unlike sentiments, these emotion scores do not have a particular range for all categories. They cannot be readily used as a feature value in the emotions category. Thus, we applied one-hot encoding to the emotions feature.

\begin{table}[!htbp]
\centering
\caption{Evaluation metrics on the test set. The top ten models comprise Random Forest (RF), Extra Trees (XTS), Decision Tree (DT), Extra Tree (XT) , Bagging (BG), NuSVC, K-Nearest Neighbors (KNN), XG Boost (XGB), Light GBM (LGBM), AdaBoost (ADB). The highest value is shown in \textbf{bold} and \textcolor{blue}{blue} color.}
\label{tbl:pred}
\begin{tabular}{|c|c|c|c|c|c|}
\hline
\textbf{Models/Metrics} & \textbf{ACC} & \textbf{PR} & \textbf{RC} & \textbf{F1} & \textbf{AUC} \\ \hline
\textbf{RF}                 & \textbf{\textcolor{blue}{0.90}}              & \textbf{\textcolor{blue}{0.90}}               & \textbf{\textcolor{blue}{0.90}}            & \textbf{\textcolor{blue}{0.90}}              & \textbf{\textcolor{blue}{0.90}}             \\ \hline
\textbf{XTS}                  & 0.90              & 0.90               & 0.89            & 0.89              & 0.90             \\ \hline
\textbf{DT}                  & 0.88              & 0.88               & 0.88            & 0.84              & 0.88             \\ \hline
\textbf{XT}                  & 0.88              & 0.88               & 0.85            & 0.87              & 0.86             \\ \hline
\textbf{BG}                  & 0.87              & 0.87               & 0.87            & 0.87              & 0.88            \\ \hline
\textbf{NuSVC}               & 0.75              & 0.76               & 0.76            & 0.76              & 0.76             \\ \hline
\textbf{XGB}                 & 0.75              & 0.75               & 0.75            & 0.77              & 0.77             \\ \hline
\textbf{KNN}                 & 0.74              & 0.73               & 0.73            & 0.73              & 0.73             \\ \hline
\textbf{LGBM}                & 0.73              & 0.73               & 0.73            & 0.73              & 0.72             \\ \hline
\textbf{ADB}                 & 0.70              & 0.70               & 0.70            & 0.70              & 0.70             \\ \hline
\end{tabular}%
\end{table}
On the train set, we use \textit{five-fold cross-validation}, which covers 80\% of the data. 
\rev{
We utilized cross-validation as implemented in Sklearn\footnote{https://scikit-learn.org/stable/modules/cross\_validation.html}. The whole dataset (114,635) is divided into train and test sets. The k-fold cross-validation has been applied to the train set to find the optimal parameters of the model, and then the test set is used to evaluate the final performance of the trained model. Following the same strategy, the 80\% train set (91,708) was used for five-fold cross-validation, and the final evaluation was performed on the 20\% test set (22,927).
}
To train and assess the model, each fold splits the train set into fold-train and fold-test sets. Finally, we evaluate the trained model's performance on an unseen test set that accounts for 20\% of the total data. Each train and test set is balanced.


The evaluation metrics for the test set are shown in Table \ref{tbl:pred} in descending order of accuracy. In our example, the ensemble-based model, Random Forest, performs best with an accuracy of \textbf{0.90}, followed by Extra Trees, Decision Tree, and so on.
This shows that writing styles may effectively distinguish between \textit{Misleading} and \textit{Non-Misleading} tweets. Other models' findings, such as Precision, Recall, F1 Score, and AUC ROC Score, show that our results are consistent throughout, showing that the trained models are generalizable.
\rev{The best performing model corresponds to a learning rate of 0.01, using 100 trees with a random state of 1.\\
}

\noindent \textbf{Feature Importance}
\begin{figure}[!htbp]
    \centering
    \includegraphics[width=9cm, height = 7cm ]{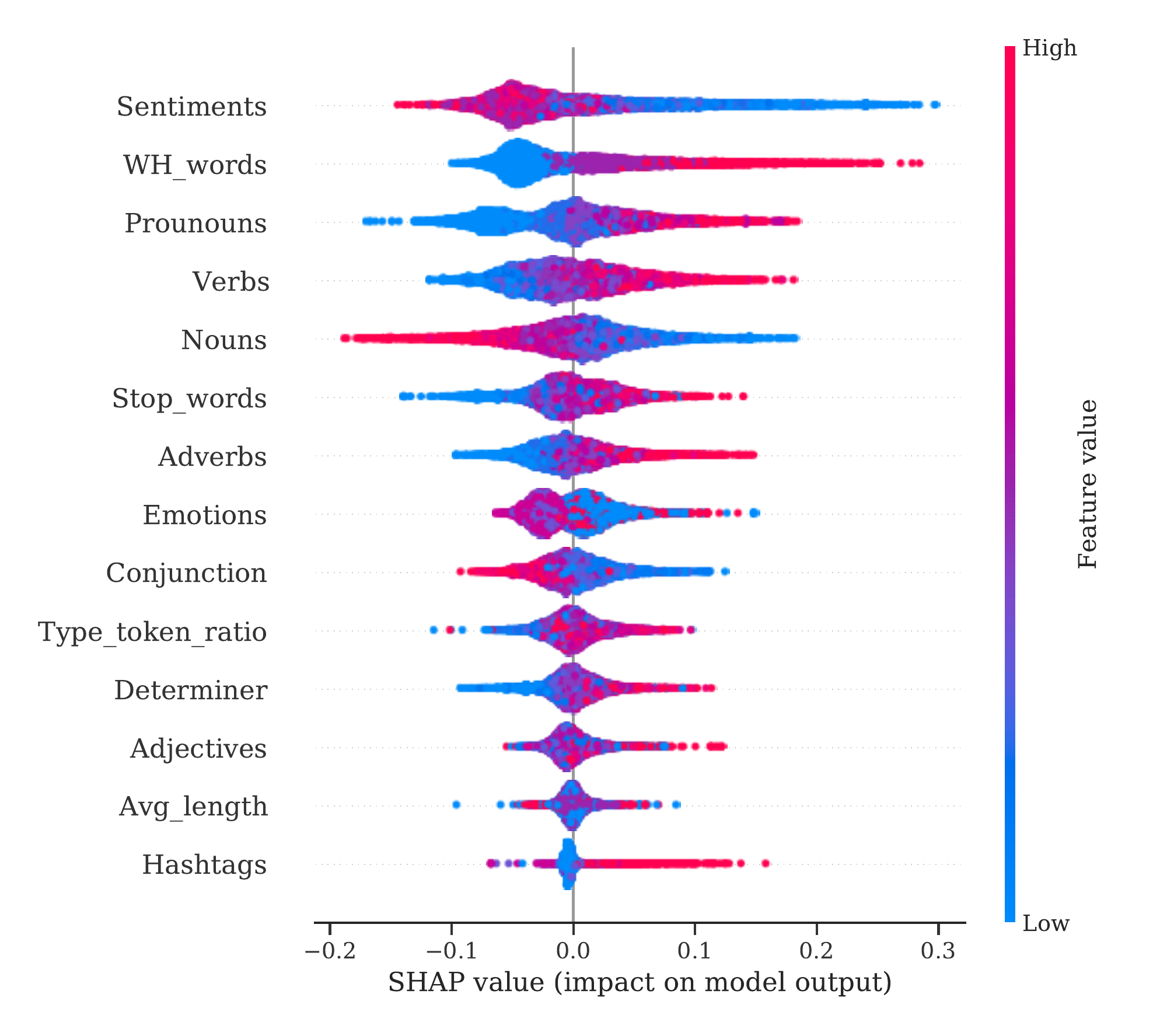}
    \caption{Feature Importance using the SHAP tool. The x-axis and y-axis denote the SHAP values and features’ names. Each data point refers to an instance of the dataset. The \textbf{\textcolor[rgb]{1, 0.2, 0.1}{red}} color indicates a higher value for the feature than its average value, whereas the \textbf{\textcolor{blue}{blue}} color denotes a lower value.\textbf{ \textcolor[rgb]{1, 0.2, 0.1}{Red}} values on the right side of the x-axis indicate a positive impact on the prediction and vice versa. Features are sorted in descending order (best seen in color).}\label{fig:shap}
\end{figure}

The SHAP Explainable AI tool\footnote{https://shap.readthedocs.io/en/latest/index.html} is then used to evaluate the contributions of each feature in the classification task.
By computing the average marginal contributions of each feature, this tool aids in finding the relevant features in the prediction. Figure \ref{fig:shap} shows the importance of the features (SHAP ranking) in descending order. \textit{Sentiments}, the most important contribution, have a negative influence on prediction, suggesting that a lower \textit{Sentiments} value predicts a \textit{Misleading} class and vice versa.
This makes sense because higher \textit{Sentiments} values indicate Positive sentiments and lower values indicate Negative sentiments, which is in line with the Section \textit{Opinion Study}'s conclusions that \textit{Misleading} tweets include more Negative sentiments. In addition, \textit{Nouns, Emotions, and Conjunctions} features have a negative influence on \textit{Misleading} tweets.
This means that compared to \textit{Non-Misleading} tweets, \textit{Misleading} tweets include fewer \textit{Nouns} and \textit{Conjunctions}. This might be because the objective of the \textit{Misleading} tweets is to entice readers by utilizing fancy words or catchy phrases rather than delivering accurate information using \textit{Nouns} and \textit{Conjunctions}.
The remaining features have a positive influence; for example, compared to \textit{Non-Misleading} tweets, \textit{Misleading} tweets have a larger number of \textit{Pronouns}.\\

\noindent \textbf{1) Feature Ablation Study:}
After evaluating the importance of features, we try to see if there is a decline in accuracy if we exclude some features. 
We experiment with a few settings, eliminating certain variables based on their importance as suggested by the SHAP ranking in Figure \ref{fig:shap}, and then rerunning all of the models in the same environment. Due to space limits, we only provide the results of the best-performing model, Random Forest. The best accuracy attained with all features is 0.90.
\begin{table}[!htbp]
\centering
\caption{Feature Ablation Study. `w/o Emo \& BF' denotes without Emotions and `Below listed Features' (BF) as per SHAP plot. Likewise, `w/o TTR \& BF', `w/o Adj \& BF' denotes without Type Token Ratio and Below listed Features, without Adjective and Below listed Features, respectively.}
\label{tbl:ablation}
\begin{tabular}{|c|c|c|c|c|c|}
\hline
\textbf{Features/Metrics}                                                                 & \textbf{ACC} & \textbf{PR} & \textbf{RC} & \textbf{F1} & \textbf{AUC} \\ \hline
\textbf{w/o Emo \& BF}        & 0.86         & 0.86        & 0.86        & 0.86        & 0.86         \\ \hline
\textbf{w/o TTR \& BF}  & 0.87         & 0.87        & 0.88        & 0.88        & 0.88         \\ \hline
\textbf{w/o Adj \& BF} & 0.88         & 0.88        & 0.88        & 0.88        & 0.89         \\ \hline
\textbf{w/o Hashtags}                                                                   & 0.89         & 0.89        & 0.89        & 0.89        & 0.89         \\ \hline
\textbf{w/ ALL features}                          & \textbf{\textcolor{blue}{0.90}}              & \textbf{\textcolor{blue}{0.90}}               & \textbf{\textcolor{blue}{0.90}}            & \textbf{\textcolor{blue}{0.90}}              & \textbf{\textcolor{blue}{0.90}}             \\ \hline
\end{tabular}
\end{table}
We start by removing \textit{Emotions} and the rest of the features listed below \textit{Emotions} as per the SHAP plot. Table \ref{tbl:ablation}, row 1 summarizes the findings. When we remove these features from our dataset, the values of the evaluation metrics decline, indicating that they are truly relevant. 
Next, we remove features that are less important than \textit{Emotions}, such as \textit{Type token ratio} and the remainder of the features (shown in Table \ref{tbl:ablation}, row 2). We continue to run experiments and discover that even the least significant feature, \textit{Hashtags}, contributes to the model's improvement. These results indicate that all features are both valuable and necessary for detecting \textit{Misleading} tweets.\\



\noindent \textbf{2) Correlation and the SHAP Ranking:}
\begin{figure*}[!htbp]
    \centering
    \includegraphics[width=15cm, height = 8cm ]{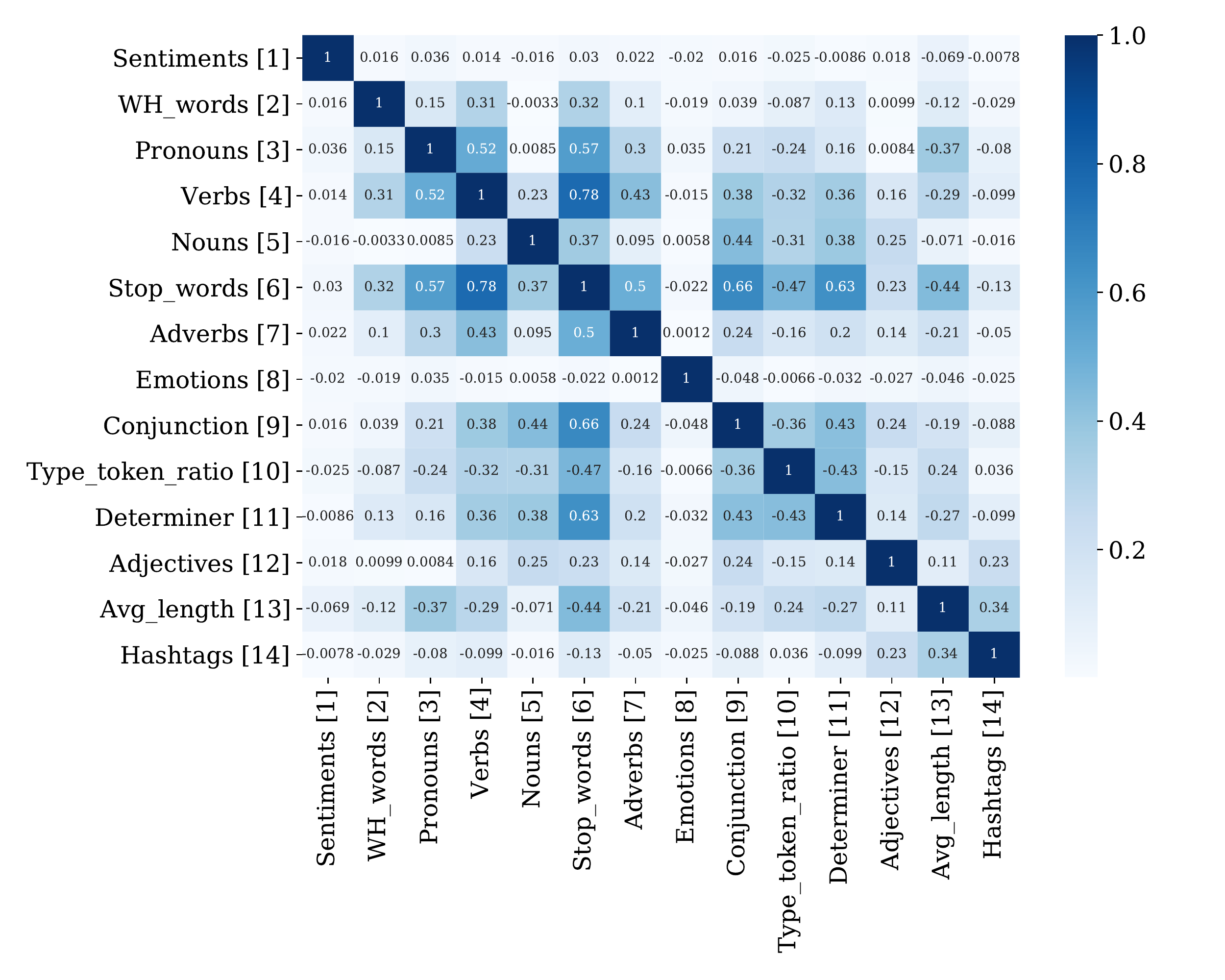}
    \caption{Correlation of the features along with SHAP ranking (represented by the numbers in the square brackets \textbf{[]}). Each cell in the symmetric matrix represents the feature pair's positive and negative correlation values. The dark color indicates a high correlation based on the absolute value. Diagonal cells are the correlation themselves, thus, showing the highest correlation.}\label{fig:corr}
\end{figure*}
Is there any association between the features' correlation values and the SHAP ranking? The hypothesis is that the highly correlated features should be close in the SHAP ranking. The correlation between each feature pair is shown in Figure \ref{fig:corr}. The dark color denotes a strong correlation between the two features based on the absolute value and vice versa. The correlation between the features does not surpass a certain threshold. This is why we use all of the features in the classification task.
The numbers in the brackets next to the feature names correspond to the feature's SHAP ranking.
We note that the highly correlated features are always positive.
Furthermore, it can be observed that highly correlated features are also close in SHAP ranking. For instance, \textit{Stop words} are highly correlated with \textit{Verbs} and score near to each other in the SHAP ranking compared to the less correlated features. \textit{Sentiments} and \textit{Determiners} is another example. \textit{Sentiments} are least correlated with \textit{Determiners} and, thus, farther from each other in SHAP ranking, demonstrating that our hypothesis is indeed true.

\section{Concluding remarks}\label{sec:conclusion}

In this paper, we carried out an exploratory analysis and meta-data associated with tweets pertaining to COVID-19 vaccines to determine the characteristics of both \textit{Misleading} and \textit{Non-Misleading} tweets. The topic detection aspect of our study helped establish the main themes of discourse across these categories, as well as identify potentially distinguishing characteristics. The latter were explored as features to carry out a classification task, where the observed outcomes support explainability.
We observe that this explainability property, coupled with the aforementioned identification of the topic of tweets, actionable intelligence can be generated, which determines a principal thrust of our future work. 

In future, we also want to explore whether the approach laid out can be generalized to identify \textit{Misleading} tweets on other topics beyond COVID-19 vaccination. \rev{The current work, given its limited scope, consequently dealt with a simple notion and dichotomy of non-misleading information. Precisely defining more complex forms of misleading information or fake news itself is a challenge, amplifying, in turn, the challenges of the task of classification.}
Beyond the extension of the work to span other topics, there is also an opportunity to refine the techniques by carrying out an analysis that is fine-grained in geographic, temporal, and linguistic dimensions.
For example, to analyze which \textit{Misleading} tweets are more prominent and specific to certain regions, which of them persist over what span of time, and doing so in languages beyond English. 

\section*{Acknowledgments}
S. Sharma and R.Sharma's work has received funding from the EU H2020 program under the
SoBigData++ project (grant agreement No. 871042), and by the CHIST-ERA grant CHIST-ERA-19-XAI-010,  ETAg (grant No. SLTAT21096).


\bibliographystyle{IEEEtran}
\bibliography{main}
\begin{IEEEbiography}[{\includegraphics[width=1in,height=1.35in,clip,keepaspectratio]{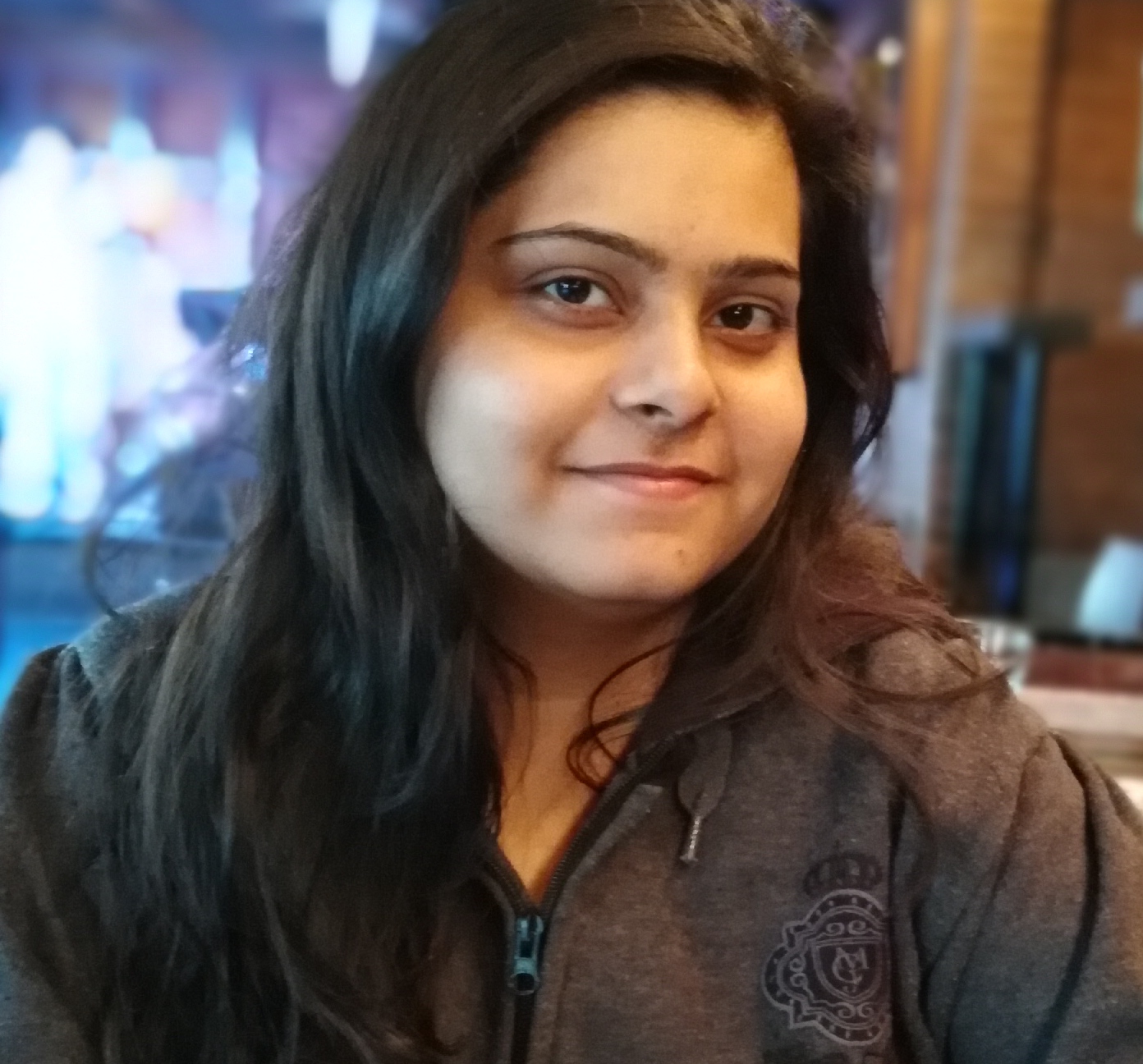}}]{Shakshi Sharma} is pursuing her PhD in the computational social science group at the Institute of Computer Science at the University of Tartu, Estonia, since June 2020.
From July 2018 to May 2020, she worked as an Assistant Professor at the Department of Computer Science and Engineering, The NorthCap University, Gurugram, Haryana, India. After earning her Master's degree from the National Institute of Technology (NIT), Delhi, India, in 2017, she also spent around a year working as a programmer at one of the Multi-National Companies (MNC): Fidelity International Company, Gurugram, Haryana, India.

Shakshi's research interests lie in the problem of Misinformation on online social media platforms. Specifically, analyzing Mis(Dis)information from various data sources and utilizing multiple AI and NLP techniques. In addition, she is working in the field of AI Ethics, focusing on the interpretability of black-box models and data biasness. 

\end{IEEEbiography}
\vskip -3\baselineskip plus -1fil
\begin{IEEEbiography}[{\includegraphics[width=1in,height=1.25in,clip,keepaspectratio]{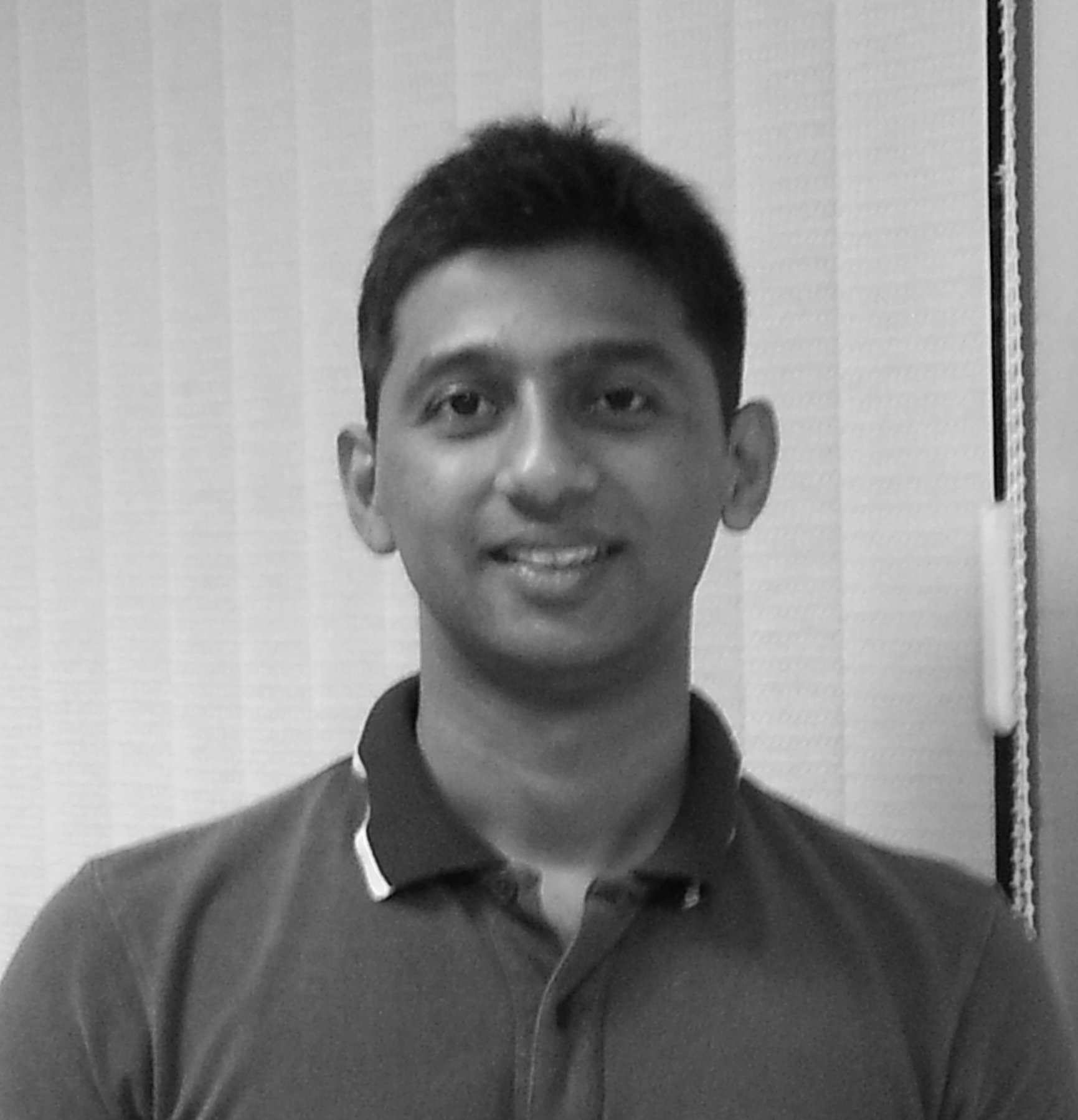}}]{Rajesh Sharma} is presently working as associate professor and leads the computational social science group at the Institute of Computer Science at the University of Tartu, Estonia, since January 2021. 

Rajesh joined the University of Tartu in August 2017 and worked as a senior researcher (equivalent to Associate Professor) till December 2020. From Jan 2014 to July 2017, he has held Research Fellow and Postdoc positions at the University of Bristol, Queen's University, Belfast, UK and the University of Bologna, Italy. Prior to that, he completed his PhD from Nanyang Technological University, Singapore, in December 2013. He has also worked in the IT industry for about 2.5 years after completing his Master's from the Indian Institute of Technology (IIT), Roorkee, India. Rajesh's research interests lie in understanding users' behavior, especially using social media data. His group often applies techniques from AI, NLP, and most importantly, network science/social network analysis. 
\end{IEEEbiography}
\vskip -2\baselineskip plus -1fil
\begin{IEEEbiography}[{\includegraphics[width=1in,height=1.35in,clip,keepaspectratio]{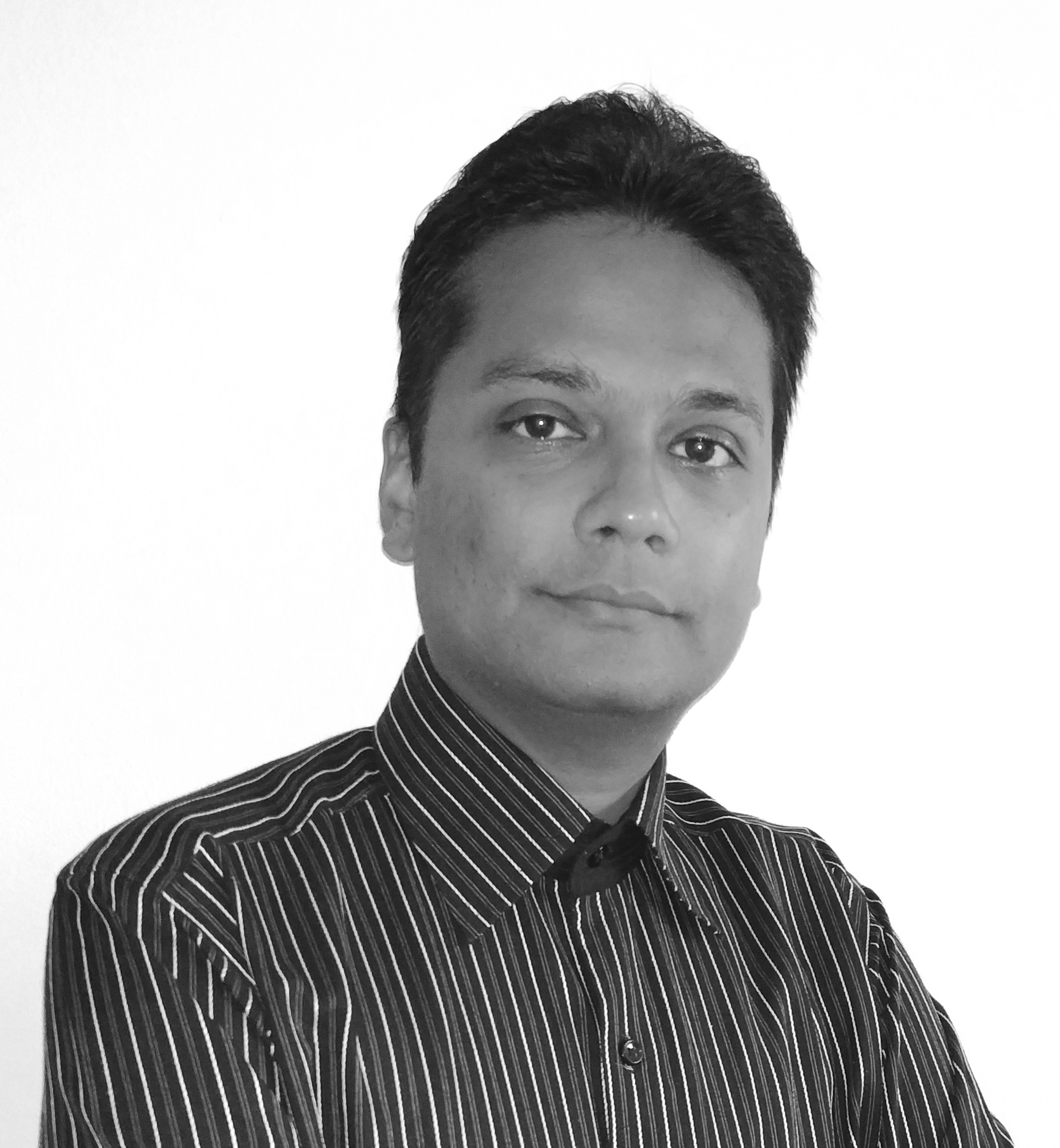}}]
{Anwitaman Datta} is an associate professor in the School of Computer Science and Engineering, Nanyang Technological University, Singapore. His core research interests span the topics of large-scale resilient distributed systems, information security and applications of data analytics. Presently, he is exploring topics at the intersection of computer science, public policies \& regulations along with the wider societal and (cyber)security impact of technology. This includes the topics of social media and network analysis, privacy, cyber-risk analysis and management, cryptocurrency forensics, the governance of disruptive technologies, as well as impact and use of disruptive technologies in digital societies and government. 
\end{IEEEbiography}

\end{document}